\documentclass[letterpaper,10 pt,conference]{ieeeconf}
\IEEEoverridecommandlockouts
\usepackage[dvipdfmx]{graphicx}
\usepackage{amsmath}
\usepackage{amssymb}
\usepackage[space]{cite}
\usepackage{bm}
\usepackage{booktabs}
\usepackage{tabularx}
\usepackage{url}
\usepackage{multirow}
\usepackage{gensymb}
\usepackage{dsfont}
\usepackage[x11names, rgb]{xcolor}
\usepackage{colortbl}
\usepackage{url}
\usepackage{siunitx}
\usepackage{threeparttable}
\usepackage{graphbox}

\usepackage{siunitx}
\usepackage{pifont}

\usepackage{tikz}
\usetikzlibrary{backgrounds}
\newcommand{\twocolorsquare}[2]{
    \begin{tikzpicture}[baseline=0.8pt,scale=0.9]
        \fill[#1] (0, 0) -- (7pt, 7pt) -- (0, 7pt) -- cycle;
        \fill[#2] (7pt, 0) -- (7pt, 7pt) -- (0, 0) -- cycle;
    \end{tikzpicture}%
}

\makeatletter
\let\NAT@parse\undefined
\makeatother
\usepackage{hyperref}
\hypersetup{
	bookmarks         = true,
	bookmarkstype     = toc,
	bookmarksnumbered = true,
	bookmarksopen     = true,
	colorlinks        = false,
	hidelinks,
	breaklinks        = false,
}

\usepackage[capitalize]{cleveref}
\crefname{section}{Sec.}{Secs.}
\Crefname{section}{Section}{Sections}
\Crefname{table}{Table}{Tables}
\crefname{table}{Tab.}{Tabs.}

\newcolumntype{C}{>{\centering\arraybackslash}X}
\newcolumntype{L}{>{\raggedright\arraybackslash}X}
\newcolumntype{R}{>{\raggedleft\arraybackslash}X}
\newcolumntype{P}[1]{>{\centering\arraybackslash}p{#1}}

\newcommand{\etal}{\textit{et al.}}

\newcommand{\s}{\hphantom{0}}

\newcommand{\myparagraph}[1]{\noindent\textbf{#1}}

\title{\LARGE \bf
Fast LiDAR Data Generation with Rectified Flows
}

\author{Kazuto Nakashima$^1$ \quad Xiaowen Liu$^2$ \quad Tomoya Miyawaki$^2$ \quad Yumi Iwashita$^3$ \quad Ryo Kurazume$^1$
\thanks{*This work was supported by JSPS KAKENHI Grant Number \{JP23K16974, JP20H00230\}.}
\thanks{$^{1}$Kazuto Nakashima and Ryo Kurazume are with the Faculty of Information Science and Electrical Engineering, Kyushu University, Japan. {\tt\small \{k\_nakashima,kurazume\}@ait.kyushu-u.ac.jp}}
\thanks{$^{2}$Xiaowen Liu and Tomoya Miyawaki are with the Graduate School of Information Science and Electrical Engineering, Kyushu University, Japan. {\tt\small \{liu,miyawaki\}@irvs.ait.kyushu-u.ac.jp}}
\thanks{$^{3}$Yumi Iwashita is with the Jet Propulsion Laboratory, California Institute of Technology, USA. {\tt\small yumi.iwashita@jpl.nasa.gov}}}

\begin{document}

\maketitle
\thispagestyle{empty}
\pagestyle{empty}

\begin{abstract}
	Building LiDAR generative models holds promise as powerful data priors for restoration, scene manipulation, and scalable simulation in autonomous mobile robots. In recent years, approaches using diffusion models have emerged, significantly improving training stability and generation quality. Despite their success, diffusion models require numerous iterations of running neural networks to generate high-quality samples, making the increasing computational cost a potential barrier for robotics applications. To address this challenge, this paper presents R2Flow, a fast and high-fidelity generative model for LiDAR data. Our method is based on rectified flows that learn straight trajectories, simulating data generation with significantly fewer sampling steps compared to diffusion models. We also propose an efficient Transformer-based model architecture for processing the image representation of LiDAR range and reflectance measurements. Our experiments on unconditional LiDAR data generation using the KITTI-360 dataset demonstrate the effectiveness of our approach in terms of both efficiency and quality.
\end{abstract}

\section{Introduction}
\label{sec:introduction}

LiDAR sensors provide accurate 3D point clouds of their surroundings using omnidirectional time-of-flight (ToF) ranging.
The LiDAR point clouds play a crucial role in enabling autonomous mobile robots to understand their surroundings both geometrically and semantically, through techniques such as SLAM, object detection, and semantic segmentation.
However, the performance of these techniques can be degraded due to incompleteness in adverse weather conditions and point density gaps between different LiDAR sensors.
Restoring the degraded point clouds requires a data prior that models complex real-world patterns.

Generative modeling of LiDAR data~\cite{caccia2019deep,nakashima2021learning,nakashima2023generative,nakashima2024lidar,zyrianov2022learning,ran2024towards,xiong2023learning,hu2024rangeldm} has been studied to address this challenge, motivated by significant progress in deep generative models~\cite{bond-taylor2022deep}.
Deep generative models aim to build neural networks that represent the probability density distribution underlying given samples.
Prior studies have demonstrated the usefulness of LiDAR generative models in tasks like sparse-to-dense completion~\cite{nakashima2021learning,nakashima2023generative,zyrianov2022learning,nakashima2024lidar} and simulation-to-real (sim2real) domain transfer~\cite{nakashima2023generative}, which also enhance perception tasks such as semantic segmentation.
Among various frameworks for the generative models, diffusion models have led to substantial improvements in the LiDAR domain~\cite{zyrianov2022learning,nakashima2024lidar,ran2024towards,hu2024rangeldm}, offering stable training and high-quality sample generation.

\begin{figure}[t]
	\centering
	\scriptsize
	\begin{tabularx}{\hsize}{CCC}
		\textbf{LiDM}~\cite{ran2024towards}$^{\dag}$ & \textbf{R2DM}~\cite{nakashima2024lidar} & \textbf{R2Flow} (ours) \\
		\cmidrule(lr){1-1} \cmidrule(lr){2-2} \cmidrule(lr){3-3}
		\begin{tikzpicture}
		\node[above right, inner sep=0] (image) at (0,0) {\includegraphics[width=\hsize]{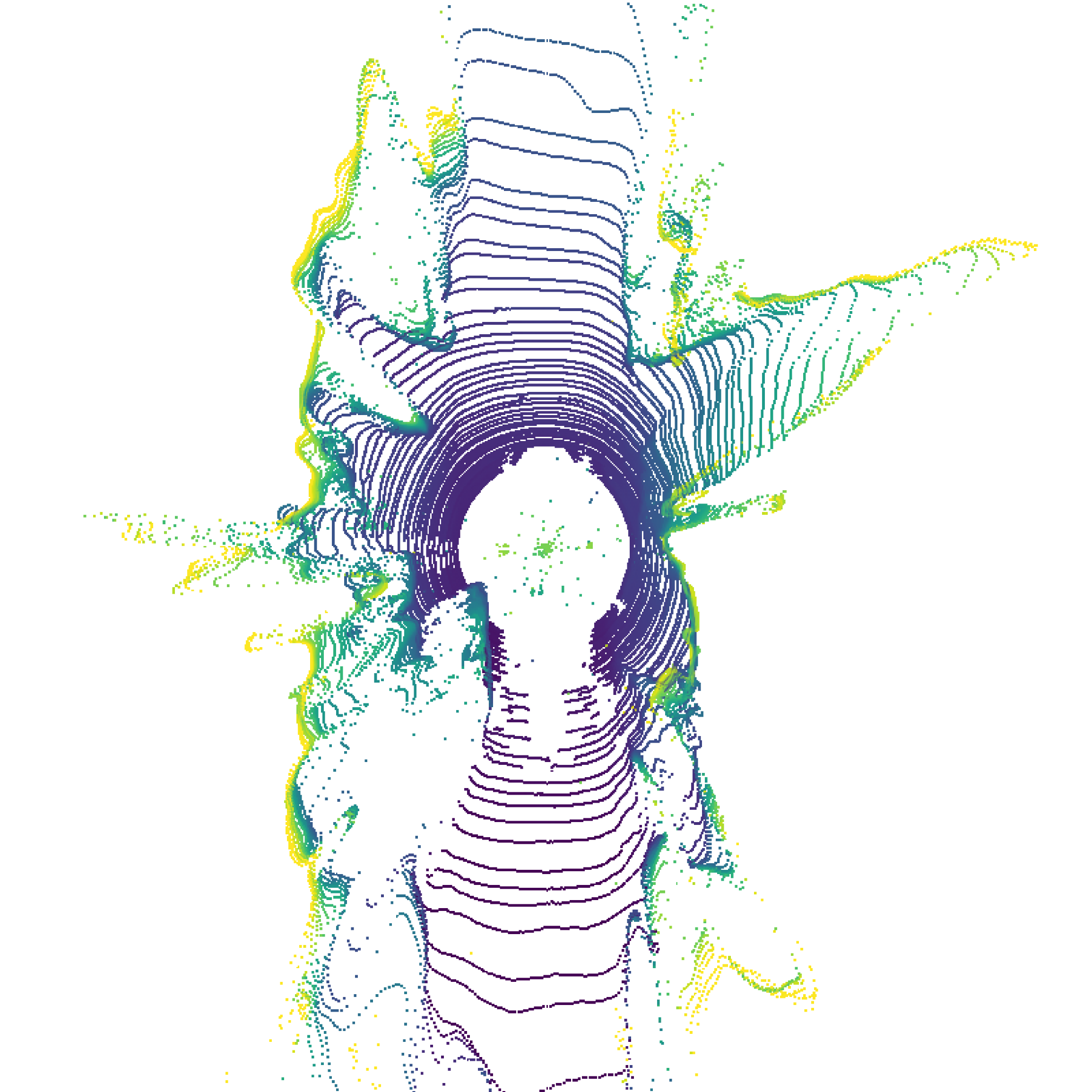}};
		\draw[thick,red] (0.5,0.2) rectangle (0.9,2.1) node[above,red,inner sep=0] (t) at (0.7,2.2) {\scriptsize blurred};
		\end{tikzpicture} & 
		\includegraphics[width=\hsize]{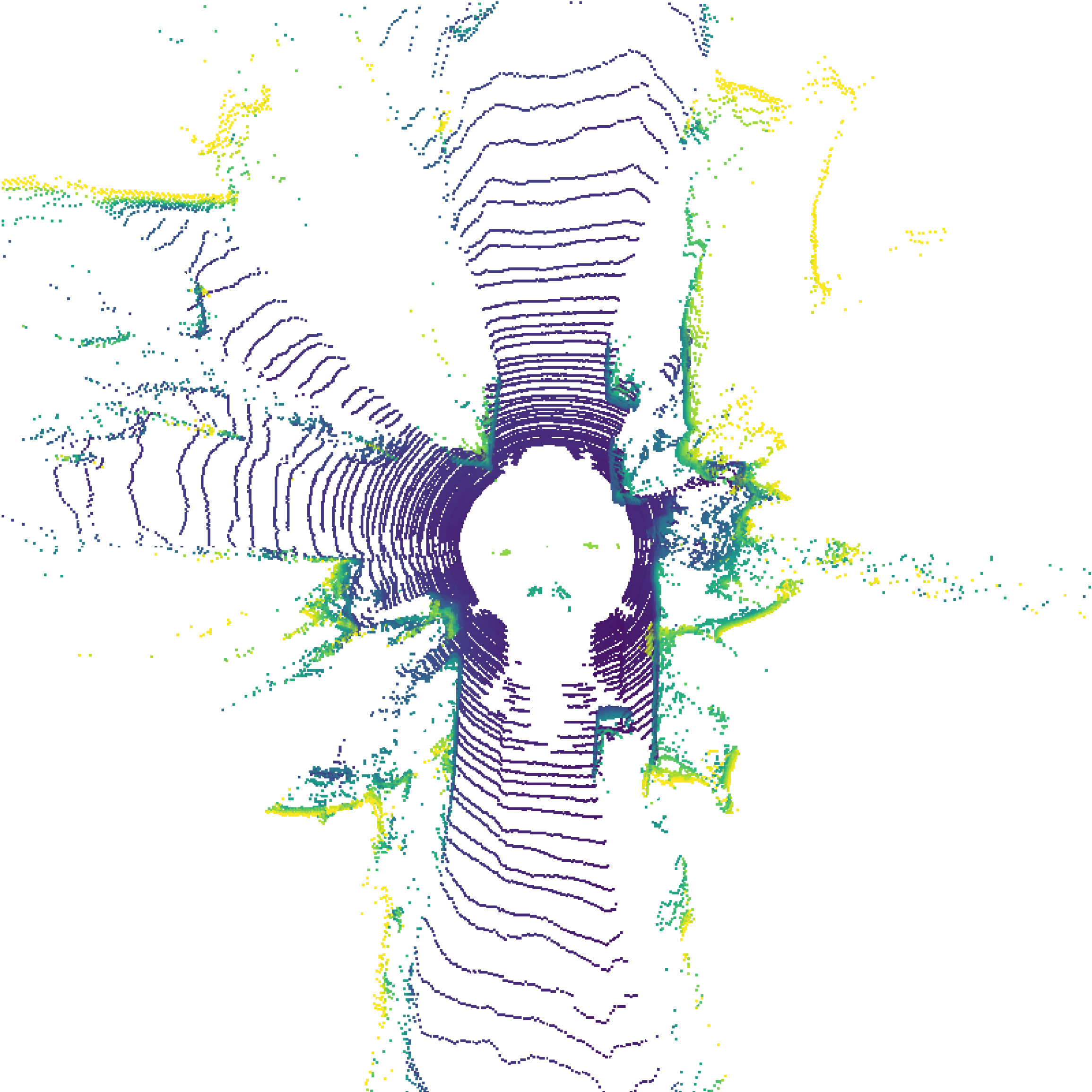} &
		\includegraphics[width=\hsize]{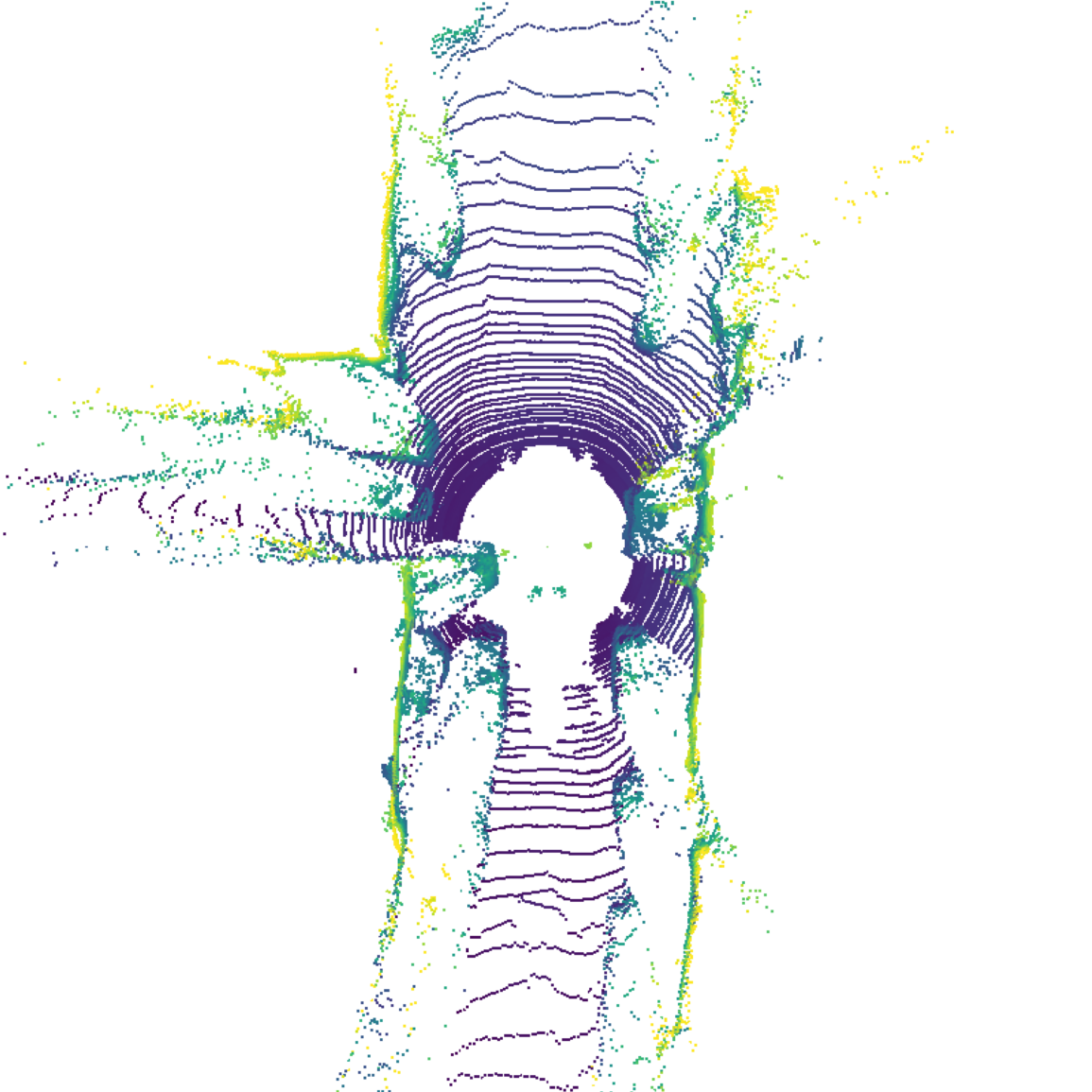} \\
		200 steps (3.7 s)                            & 256 steps (3.7 s)                       & 256 steps (7.6 s)      \\
		\begin{tikzpicture}
		\node[above right, inner sep=0] (image) at (0,0) {\includegraphics[width=\hsize]{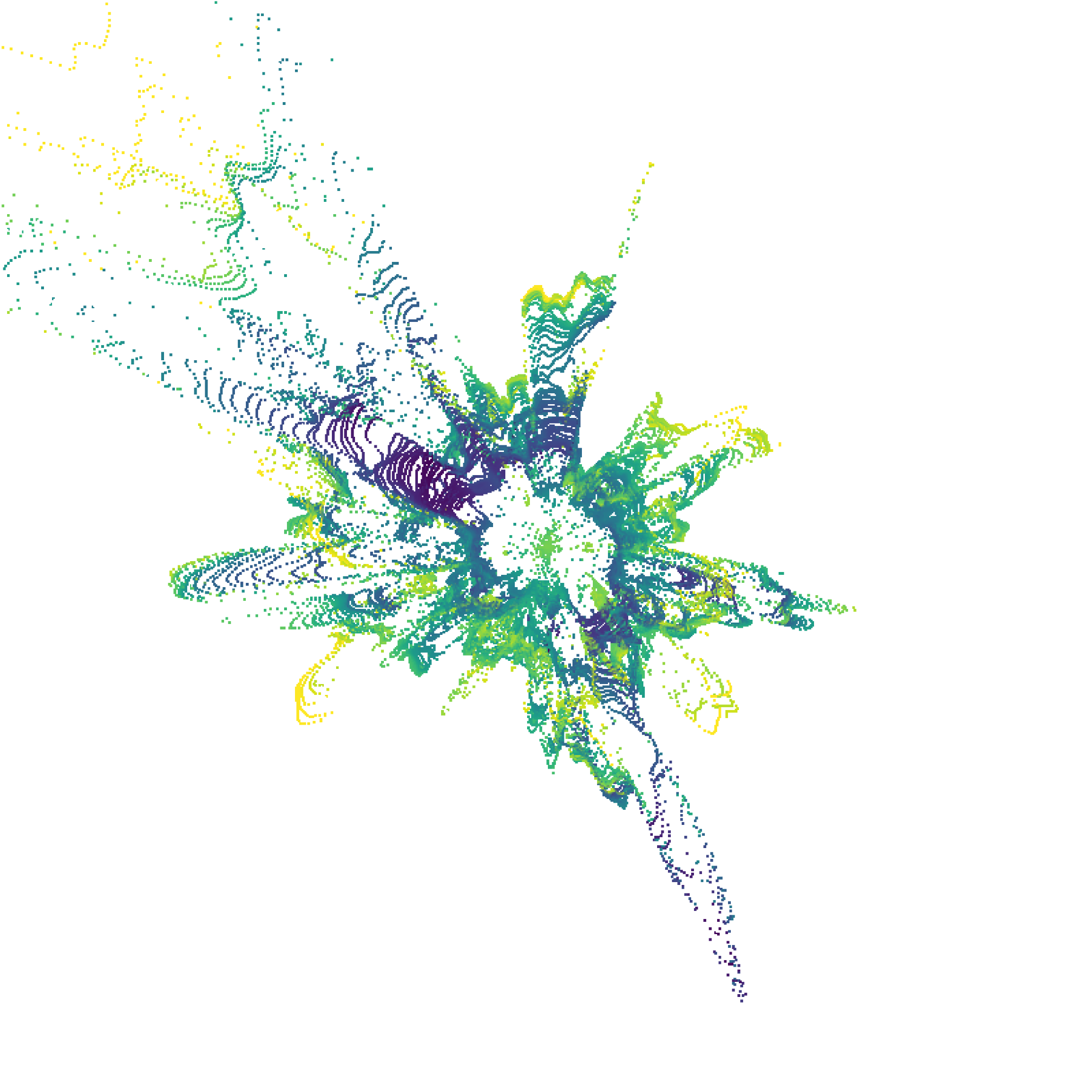}};
		\node[above right,red,inner sep=0] (t) at (0.3,0.2) {\scriptsize corrupted};
		\end{tikzpicture} & 
		\begin{tikzpicture}
		\node[above right, inner sep=0] (image) at (0,0) {\includegraphics[width=\hsize]{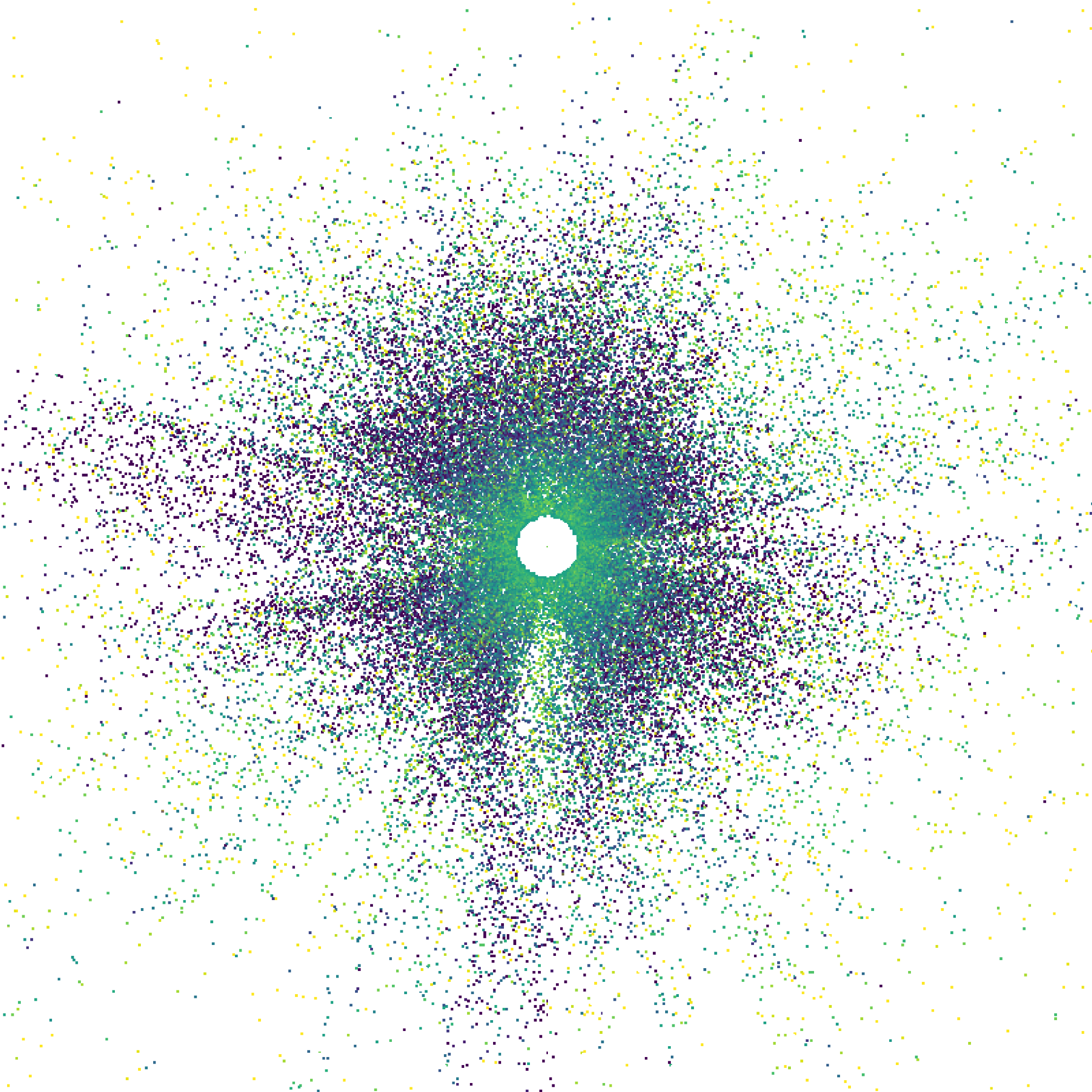}};
		\node[above right,red,inner sep=0] (t) at (0.3,0.2) {\scriptsize corrupted};
		\end{tikzpicture} &
		\includegraphics[width=\hsize]{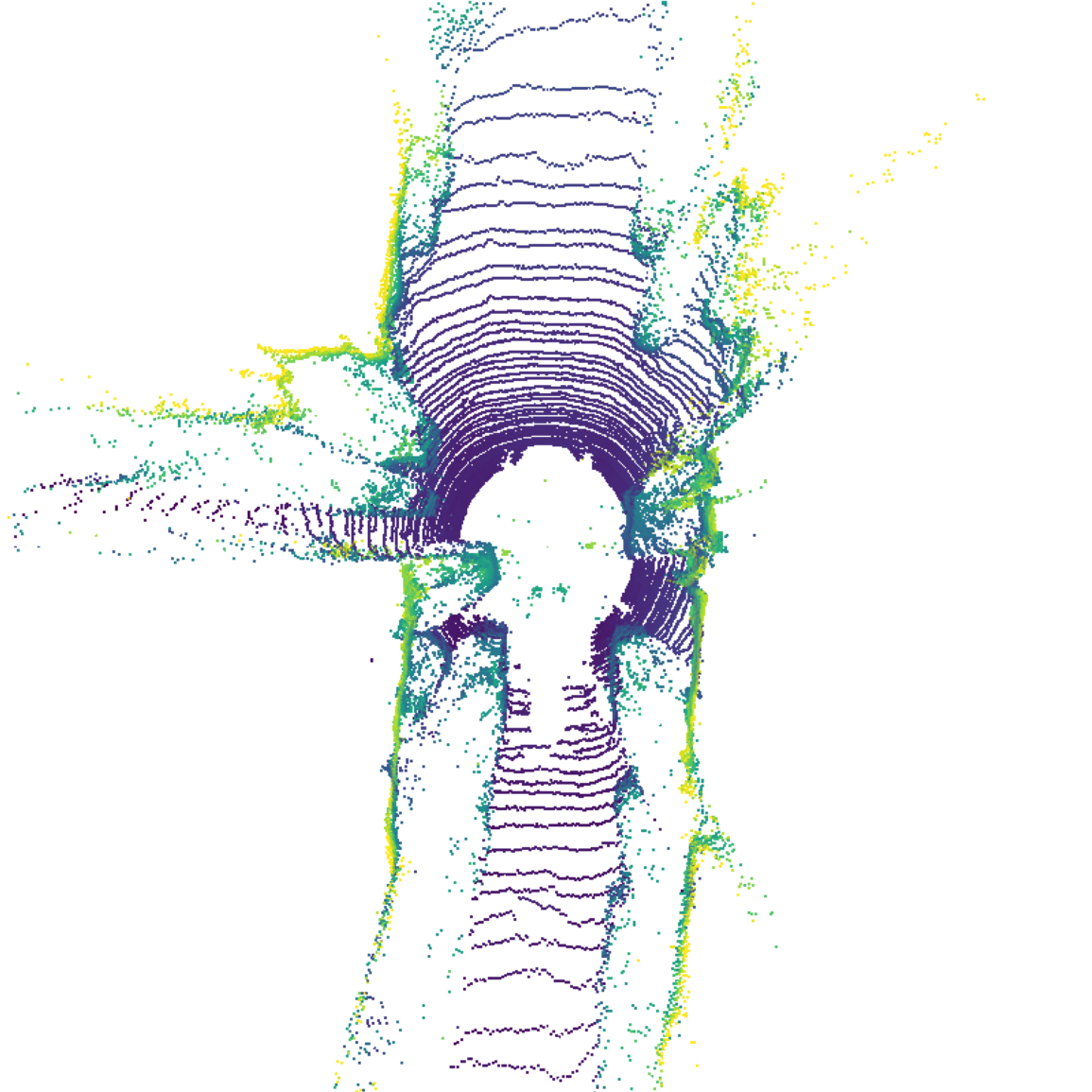}  \\
		2 steps (38 ms)                              & 2 steps (56 ms)                         & 2 steps (62 ms)        
	\end{tabularx}
	\caption{\textbf{Comparison of LiDAR generative models}. Diffusion models have demonstrated realistic LiDAR data generation, while the previous methods~\cite{ran2024towards,nakashima2024lidar} suffer from the trade-off between quality and sampling efficiency in their \textit{iterative} generation process. Our approach consistently generates high-quality samples across different numbers of iterations. $\dag$ Our improved version with APE~\cite{peebles2023scalable}.}
	\label{fig:teaser}
\end{figure}

Despite their success, diffusion models require significant computational costs to generate high-quality LiDAR data.
In general, sampling in diffusion models is formulated as a stochastic differential equation (SDE)~\cite{song2021score-based} that describes the sample trajectories from a latent distribution to the data distribution.
Accurately simulating these learned trajectories requires hundreds or even thousands of discretized steps, each involving the execution of a deep neural network.
\cref{fig:teaser} illustrates the trade-offs between the number of sampling steps and the sample quality for the latest methods on the LiDAR data generation~\cite{ran2024towards,nakashima2024lidar}.
As shown in~\cref{fig:teaser}, naively reducing the number of sampling steps degrades the quality of the generated LiDAR data.
This limitation poses a challenge for robotics applications, where power efficiency and computational speed are critical constraints.

To this end, we propose \textbf{R2Flow} (\textbf{R}ange–\textbf{R}eflectance \textbf{Flow}), a novel generative model for fast and realistic LiDAR data generation.
As a framework for building generative models, we employ rectified flows~\cite{liu2023flow,lee2024improving}, a type of conditional flow matching framework designed to train continuous normalizing flows~\cite{lipman2023flow,tong2024improving}.
Rectified flows have been successfully applied in natural image generation tasks, such as human faces and common objects.
Similar to diffusion models, the rectified flows represent the data generation as an iterative transformation using a deep neural network.
However, a key distinction is that rectified flows use \textit{deterministic straight} trajectories, whereas diffusion models employ \textit{stochastic curved} trajectories.
This makes sampling robust to the number of steps or step size, with the potential to simulate the entire trajectory in just a single step.
Following relevant studies~\cite{caccia2019deep,nakashima2021learning,nakashima2023generative,nakashima2024lidar,zyrianov2022learning,ran2024towards}, our R2Flow is trained on the equirectangular image representation of multimodal measurements: range and reflectance (intensity of laser reflection).
We also propose a neural network architecture to generate the multimodal images based on the lightweight Vision Transformer (ViT)~\cite{crowson2024scalable}.
Among recent architectures, we verify that our approach achieves better sample quality for LiDAR data generation while reducing the computational cost and the model size.
We evaluate our approach through an unconditional generation task on the KITTI-360 dataset~\cite{liao2022kitti-360}.
Our approach outperforms the state-of-the-art results for both large and small numbers of steps.
We summarize our contributions as follows:
\begin{itemize}
	\item We propose R2Flow, a rectified flow-based deep generative model for fast and realistic generation of LiDAR range and reflectance modalities.
	\item We introduce a ViT-based model architecture that balances fidelity and efficiency in LiDAR data generation.
	\item We demonstrate the effectiveness of our approach through an unconditional generation task on the KITTI-360 dataset.
\end{itemize}

\section{Related Work}
\label{sec:related_work}

Here, we briefly summarize three classes of existing LiDAR generative models using neural networks.

\myparagraph{Variational autoencoders (VAEs).}
VAEs~\cite{kingma2014auto-encoding} are trained with an autoencoder with latent representation at the bottleneck.
Caccia~\etal~\cite{caccia2019deep} initiated early work on generating LiDAR range images using a vanilla VAE.
While VAEs provide stable training with the ELBO objective, they often produce blurry samples.
More recently, Xiong~\etal~\cite{xiong2023learning} employed the improved framework, a vector-quantized variational autoencoder (VQ-VAE)~\cite{vandenoord2017neural}, with the voxel-based LiDAR data representation. 

\myparagraph{Generative adversarial networks (GANs).}
GANs~\cite{goodfellow2014generative} consist of two competing neural networks, a generator and a discriminator, and have been actively applied in various domains over the last decade. Caccia~\etal~\cite{caccia2019deep} reported the first results by training a basic GAN on range images.
DUSty~\cite{nakashima2021learning} and DUSty v2~\cite{nakashima2023generative} proposed architectures designed to be robust against raydrop noise (missing points caused by non-returned laser signals). Although GANs achieve better sample quality than VAEs, they suffer from unstable training and generated point clouds still deviate from real samples.

\begin{figure}[t]
	\centering
	\scriptsize
	\begin{tabularx}{\hsize}{CC}
		\includegraphics[width=0.9\hsize]{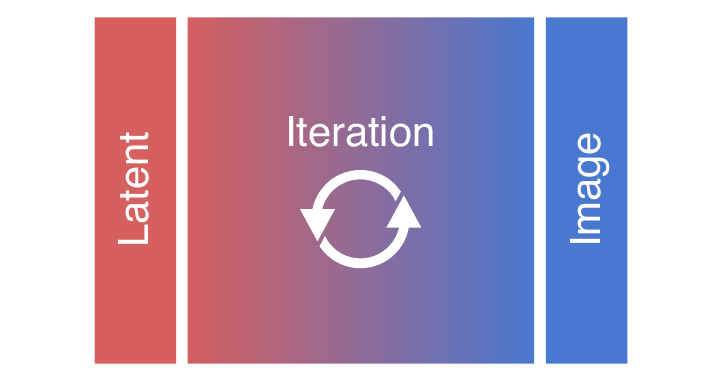}                       &                                  
		\includegraphics[width=0.9\hsize]{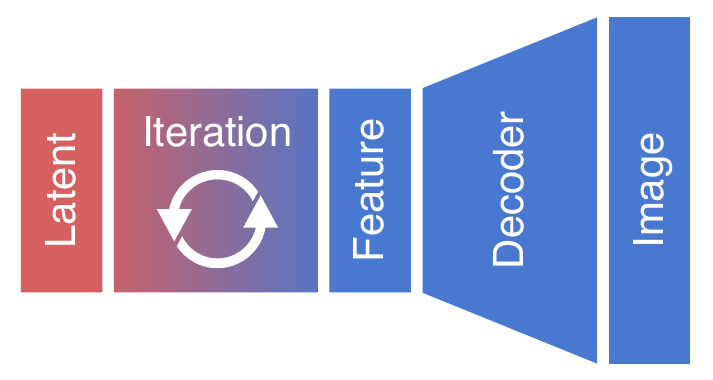} \\
		\textbf{Pixel-space iteration}                                                          & \textbf{Feature-space iteration} \\
		(LiDARGen~\cite{zyrianov2022learning}, R2DM~\cite{nakashima2024lidar}, \textbf{R2Flow}) &                                  
		(LiDM~\cite{ran2024towards}, RangeLDM~\cite{hu2024rangeldm}) \\
	\end{tabularx}
	\caption{\textbf{Architectural comparison of LiDAR diffusion models and ours}. Our approach R2Flow is categorized into the pixel-space iteration approach.}
	\label{fig:architectural_comparison}
\end{figure}

\myparagraph{Diffusion models.}
In recent years, diffusion models have gained significant attention for their stable training and high-quality sample generation.
Diffusion models define bidirectional transitions based on a multi-step Markov process between data and latent variable spaces of the same dimensionality.
Various formulations, such as score matching with Langevin dynamics (SMLD)~\cite{song2019generative,song2020improved} and denoising diffusion probabilistic modeling (DDPM)~\cite{ho2020denoising,kingma2021variational}, have been proposed to schedule these transitions.
It is also known that these formulations can be generalized as stochastic differential equations (SDEs)~\cite{song2021score-based}.
Several studies have applied diffusion models to LiDAR data generation.
LiDARGen~\cite{zyrianov2022learning} employs SMLD, also known as a variance exploding SDE~\cite{song2021score-based}, to train range and reflectance images in pixel-space using a discrete-time schedule.
R2DM~\cite{nakashima2024lidar} employs DDPM, also known as a variance preserving SDE~\cite{song2021score-based}, to also train range and reflectance images in pixel-space using a continuous-time schedule.
Pixel-space diffusion models can capture fine details, but they incur high computational costs due to the iterative nature of sampling.
To mitigate this issue, Ran~\etal~\cite{ran2024towards} proposed LiDM with architectural improvements based on the latent diffusion model (LDM)~\cite{rombach2022high-resolution}.
LiDM first pre-trained an autoencoder to compress the range images and then trained a discrete-time diffusion model on the lower dimensional feature space.
RangeLDM~\cite{hu2024rangeldm} is a more recent work following a similar LDM approach.
Nevertheless, the LDM approaches still struggle with blurriness caused by the non-iterative decoding by the autoencoder.
\cref{fig:architectural_comparison} illustrates architectural comparison of the pixel-space and feature-space approaches.
We prioritize the pixel precision required for range images and employ the pixel-space approach.

In summary, the diffusion model-based approach can generate high-fidelity samples with stable training among the generative model frameworks.
However, the sampling process requires a sufficiently large number of steps because the generative process is defined by stochastic curved trajectories defined as SDEs.
If these trajectories are approximated with too few steps, the generated LiDAR samples are prone to discretization errors (see \cref{fig:teaser} for an example).
In this paper, we address this issue by introducing \textit{easy-to-approximate} trajectories demonstrated in natural image domains~\cite{liu2023flow}.

\section{Method}
\label{sec:approach}

We employ rectified flow~\cite{liu2023flow} and its extension~\cite{lee2024improving} to construct generative trajectories optimized for straightness, enabling efficient sampling with only a few steps. In this section, we first introduce the procedure for building straight trajectories by rectified flows and then describe our modifications to LiDAR data generation.

\subsection{Preliminary}
\label{sec:preliminary}

\myparagraph{Initial training.}
Suppose the unknown data distribution $p_1$ where the dataset samples $\bm{x}_1\sim p_1$ are only accessible, and Gaussian distribution $p_0=\mathcal{N}(0,\bm{I})$ which draws the latent variables $\bm{x}_0\sim p_0$.
Both $p_1$ and $p_0$ are defined over $\mathbb{R}^{d}$.
The goal is to build a transport map between the two distributions.
In rectified flows, the data transformation is formulated as the following ordinary differential equation (ODE):
\begin{align}
	\label{eq:ode}                                 
	d\bm{x}_t=v_{\theta}\left(\bm{x}_t,t\right) dt 
\end{align}
where $\bm{x}_t\in\mathbb{R}^{d}$ is an intermediate state at timestep $t\in[0,1]$ and $v_{\theta}: \mathbb{R}^{d}\to\mathbb{R}^{d}$ is a neural network to predict the velocity fields towards $\bm{x}_1$.
To minimize the discretization errors in ODE integration, a linear interpolation path is considered: $\bm{x}_t = t \bm{x}_1 + (1-t) \bm{x}_0$.
\cref{fig:model}(a) illustrates the example trajectory.
Then we train the neural network $v_{\theta}$ by minimizing the following conditional flow matching loss $\mathcal{L}_{\rm{CFM}}$ for independently sampled pairs $(\bm{x}_1, \bm{x}_0)$, so that $v_{\theta}$ encourages the sample $\bm{x}_t$ to follow the uniform velocity $\bm{x}_1-\bm{x}_0$ as closely as possible~\cite{liu2023flow}.
\begin{align}
	\label{eq:flow_matching_loss}                                                                                               
	\mathcal{L}_{\rm{CFM}}=\mathbb{E}\left[\| \left(\bm{x}_1 - \bm{x}_0\right) - v_\theta\left(\bm{x}_t, t\right)\|^2_2\right], 
\end{align}
where $t\sim \mathrm{Uniform}(0,1)$.
The initial model $v_{\theta}$ obtained from~\cref{eq:flow_matching_loss} is referred to as 1-RF in this paper.

\myparagraph{Straightening.}
Although the initial model 1-RF is capable of producing high-quality samples, the built trajectories are not straight because the model is trained on independently sampled pairs $(\bm{x}_1, \bm{x}_0)$.
As a result, 1-RF still requires a large number of steps for sampling.
Rectified flows address this issue by iteratively refining the flow field through \textit{reflow}~\cite{liu2023flow}.
In the reflow process, $\bm{x}_0$ is sampled from $p_0$, while the target point $\bm{x}_1$ is obtained by solving \cref{eq:ode}, using 1-RF and the initial value $\bm{x}_0$.
Training $v_{\theta}$ with the newly dependent pairs $(\bm{x}_1, \bm{x}_0)$ reduces the transport cost between $p_0$ and $p_1$, leading the sample trajectories to become straighter~\cite{liu2023flow}.
Following the improved technique proposed by Lee~\etal~\cite{lee2024improving}, we switch the loss to the following pseudo-Huber loss:
\begin{align}
	\label{eq:reflow_huber_loss}                                                                                                             
	\mathcal{L}_{\rm{PH}}= \mathbb{E}\left[\sqrt{\| \left(\bm{x}_1 - \bm{x}_0\right) - v_\theta\left(\bm{x}_t, t\right)\|^2_2+c^2}-c\right], 
\end{align}
where $c=0.00054\sqrt{d}$ and $d$ is the dimension of the data.
It is known that the 1-RF loss is difficult to be minimized around $t=0$ and $t=1$~\cite{lee2024improving}.
We will also observe the same phenomenon in~\cref{fig:straightness}.
To give more weights for the timesteps, we sample $t$ from the U-shaped distribution~\cite{lee2024improving}: $p_t(u) \propto e^{au} + e^{-au}$ where $u\in[0,1]$ and $a=4$.
The model obtained by reflow is called 2-RF in this paper.

\myparagraph{Timestep distillation.}
The straightened 2-RF model can be further improved by timestep distillation~\cite{liu2023flow}.
At this stage, the model training focuses on the specific timesteps required for few-step sampling, sacrificing predictions at the other unnecessary timesteps.
For instance, distilling to a 2-step sampling involves training the model only at $t\in\{0,0.5\}$.
All other settings remain the same as in the 2-RF.
We denote the $i$-RF model distilled with $k$-step as $i$-RF + $k$-TD.

\myparagraph{Sampling.}
Sampling can be performed by solving the initial value problem of the ODE described in \cref{eq:ode}, using the learned velocity estimators $v_{\theta}$.
For 1-RF and 2-RF, we can choose the arbitrary number of sampling steps.
In general, the larger the number of steps, the smaller the discretization error.
For $k$-TD models, the number of sampling steps is fixed at $k$.
Any integration solvers can be used for sampling, such as the following Euler method:
\begin{align}
	\label{eq:euler_sampler}                                                                                     
	\bm{x}_{t_{n+1}} \leftarrow \bm{x}_{t_n} + \left(t_{n+1}-t_n\right) v_{\theta}\left(\bm{x}_{t_n},t_n\right), 
\end{align}
where $0 \leq t_n < t_{n+1} < 1$ and $\bm{x}_0\sim N(0,\bm{I})$.

\begin{figure}[t]
	\centering
	\scriptsize
	\includegraphics[width=\hsize]{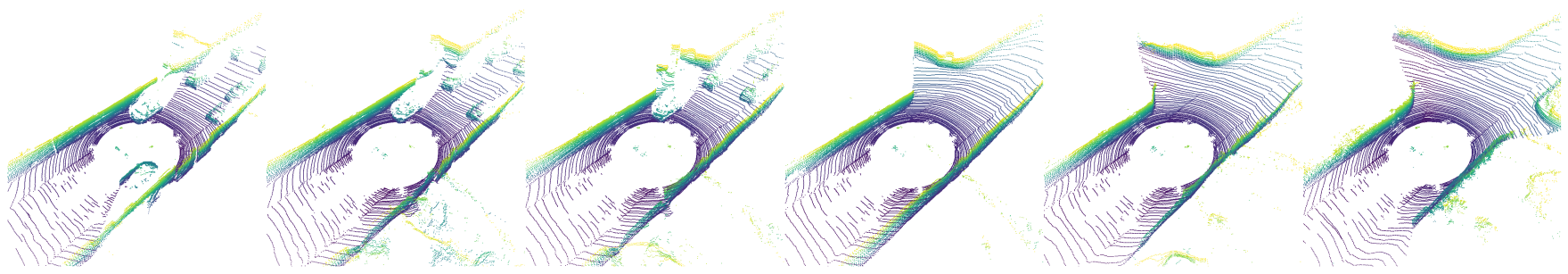}
	\begin{tabularx}{\hsize}{CCCCCC}
		\cmidrule(lr){1-1} \cmidrule(lr){2-5} \cmidrule(lr){6-6}
		Inversion & \multicolumn{4}{c}{Spherical linear interpolation (slerp) on latent space} & Inversion \\
		  &   &   &   &   &   \\ 
	\end{tabularx}
	\vspace{-5mm}
	\caption{\textbf{Scene interpolation using R2Flow inversion.} The both side were reconstructed from real samples via inversion. The middle four samples were generated using interpolated latent variables.}
	\label{fig:interpolation}
\end{figure}

\myparagraph{Inversion.}
The rectified flows can perform \textit{inversion}, a process of transforming given data $\bm{x}_1$ into the corresponding embedding $\bm{x}_0$ in the latent space, by solving ODE in Eq.~\ref{eq:ode} reversely, from $t=1$ to $t=0$.
Similar to the other generative models, the inverted $\bm{x}_0$ can be used for various applications such as image manipulation.
We showcase the application of LiDAR scene interpolation using our model in~\cref{fig:interpolation}.

\subsection{Data Representation}

Following the existing studies~\cite{caccia2019deep,nakashima2021learning,nakashima2023generative,nakashima2024lidar,zyrianov2022learning,ran2024towards}, R2Flow is trained on the equirectangular image representation of LiDAR data.
We assume a LiDAR sensor that has an angular resolution of $W$ in azimuth and $H$ in elevation and measures the range and reflectance at each laser angle.
Then, $HW$ sets of the range and reflectance values can be projected to a 2-channel equirectangular image $\bm{x}_1\in\mathbb{R}^{2\times{H}\times{W}}$ by spherical projection.
Moreover, Following prior work~\cite{zyrianov2022learning,nakashima2024lidar}, we also rescale the range modality $\bm{x}_{\rm{range}}\in[0, x_{\rm{max}}]^{1\times{H}\times{W}}$ to a log-scale representation $\bm{x}_{\rm{log}}\in[0, 1]^{1\times{H}\times{W}}$ as follows:
\begin{equation}
	\bm{x}_{\rm{log}}=\frac{\mathrm{log}(\bm{x}_{\rm{range}}+1)}{\mathrm{log}(x_{\mathrm{max}}+1)}.
\end{equation}
This log-scale representation gains the geometric resolution of nearby points.
Generated range images can be projected back to the 3D point clouds with the reflectance values.

\subsection{Velocity Estimator}

\begin{figure*}[t]
	\centering
	\footnotesize
	\includegraphics[width=\hsize]{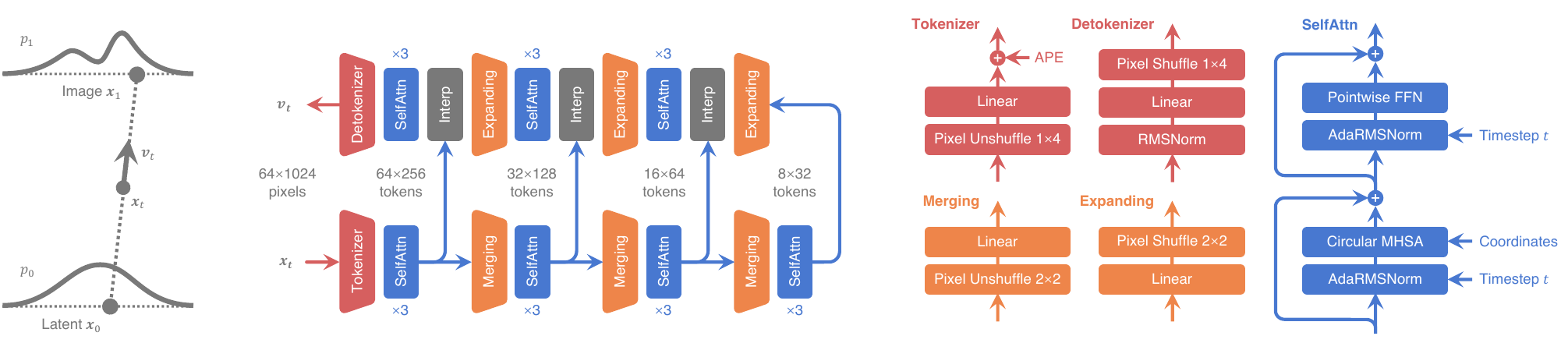}
	\begin{tabularx}{\hsize}{P{20mm}P{75mm}P{55mm}}
		(a) Straight flows                                             &   
		(b) HDiT~\cite{crowson2024scalable}-based overall architecture &   
		(c) Details of the building blocks
	\end{tabularx}
	\caption{\textbf{Schematic overview of our velocity estimator}. (a) Straight flows are learned to transport samples between the latent space $p_0$ and the image space $p_1$. (b) Overall architecture to estimate the velocity fields $\bm{v}_t$ from the intermediate state $\bm{x}_t$ and the timestep $t$. The \texttt{Interp} layers fuse the current tokens and skipped tokens at each spatial location with learnable weights. (c) The details of the building blocks. The \texttt{Circular MHSA} (multi-head self-attention) layer uses a global attention kernel at the bottleneck and a sliding local window~\cite{hassani2023neighborhood} for other stages.}
	\label{fig:model}
\end{figure*}

In this section, we describe the design choice of the velocity estimator $v_{\theta}$ in~\cref{eq:ode}.
\cref{fig:model} depicts the schematic diagram of the model architecture.

\myparagraph{Pixel-space vs. feature-space.}
As discussed in~\cref{sec:related_work}, the pixel-space iteration involves high computational cost in general, which poses a barrier to the adoption of powerful backbone models such as Vision Transformers (ViTs)~\cite{dosovitskiy2021image}.
A common approach to this issue is to reduce the dimensionality at which iterative models operate. 
In the context of diffusion models, Rombach \etal~\cite{rombach2022high-resolution} proposed LDM (latent diffusion model), which consists of an autoencoder (AE) pretrained to \textit{perceptually} compress images and a diffusion model trained on the lower-dimensional AE features. 
The compression is motivated by the observation that representing imperceptible details of natural images can be relegated to AE.
However, the AE-based non-iterative decoding can be problematic in LiDAR range image generation which requires accurate pixel values and their alignment to maintain geometric fidelity in point clouds. 
For instance, LiDM~\cite{ran2024towards} proposed the LDM-based approach in the LiDAR generation task but also identified blurry patterns output by the AE decoder as a remaining issue (see \cref{fig:teaser,fig:generated_samples}).
Therefore, in this paper, we reconsider powerful yet efficient pixel-space architectures for precise modeling, while minimizing the number of iterations by rectified flows.

\myparagraph{Architecture design.}
Our model is built upon HDiT (hourglass diffusion transformer)~\cite{crowson2024scalable}, which is a ViT-based architecture proposed for pixel-space diffusion models.
The key idea is to use a sliding window self-attention mechanism~\cite{hassani2023neighborhood} to avoid increasing the $\mathcal{O}(n^2)$ computation with respect to the number of tokens $n$.
It thereby enables high-resolution image generation without requiring the AE-based compression~\cite{rombach2022high-resolution} or staged upsampling~\cite{saharia2022photorealistic}, despite the pure Transformer structure.
To adapt HDiT for \textit{panoramic} LiDAR range and reflectance image generation, we introduce the following modifications.
(i) The sliding window in the self-attention layers~\cite{hassani2023neighborhood} is modified to operate in a horizontal circular pattern using a circular padding technique~\cite{zyrianov2022learning,nakashima2024lidar}.
(ii) Following the recent ViT-based architecture for LiDAR processing~\cite{yang2024tulip}, the patch size in tokenization is changed from the default square shape to a landscape shape of $1\times4$. The sliding windows also has the landscape shape of $3\times9$.
(iii) We use pre-defined LiDAR beam angles to condition the \textit{relative} positional embeddings (RoPE~\cite{su2021roformer}) in the self-attention layers, limiting the angular frequencies to harmonics.
(iv) Similar to ViT~\cite{dosovitskiy2021image}, we apply a learnable additive bias to the tokens as an \textit{absolute} positional embedding (APE); otherwise, the generated LiDAR point clouds involve random azimuth rotation.

\section{Experiments}
\label{sec:experiments}

In this section, we present the quantitative and qualitative evaluation of the unconditional generation task, focusing on the faithfulness of the sampled LiDAR data.

\subsection{Settings}

\myparagraph{Dataset.}
Following prior work~\cite{zyrianov2022learning,ran2024towards,nakashima2024lidar}, we utilize the KITTI-360~\cite{liao2022kitti-360} dataset. The KITTI-360 dataset contains 81,106 point clouds captured using a Velodyne HDL-64E (64-beam mechanical LiDAR sensor). We adopt the standard data split defined by Zyrianov~\etal~\cite{zyrianov2022learning}.
Each point cloud is projected onto a $64\times1024$ image with range and reflectance values assigned to each pixel.

\myparagraph{Baselines.}
We selected baseline methods for which implementations are publicly available. For GAN-based approaches, we compare the vanilla GAN~\cite{caccia2019deep} (stable version in~\cite{nakashima2021learning}), DUSty v1~\cite{nakashima2021learning}, and DUSty v2~\cite{nakashima2023generative}. For diffusion-based approaches, we include comparisons with LiDARGen~\cite{zyrianov2022learning}, R2DM~\cite{nakashima2024lidar}, and LiDM~\cite{ran2024towards}.
We re-trained the GAN models on the KITTI-360 dataset both with and without the reflectance modality. For LiDARGen and R2DM, we used the available pre-trained weights to generate samples. Additionally, we trained LiDM (excluding the autoencoder part) using the same training split, alongside an improved model described in the following.

\begin{figure}[t]
	\centering
	\scriptsize
	\setlength{\tabcolsep}{2pt}
	\begin{tabularx}{\hsize}{CCC}
		\includegraphics[width=0.9\hsize]{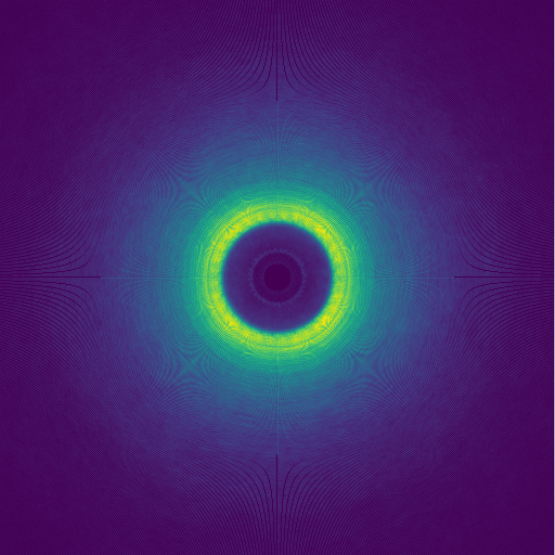} &
		\includegraphics[width=0.9\hsize]{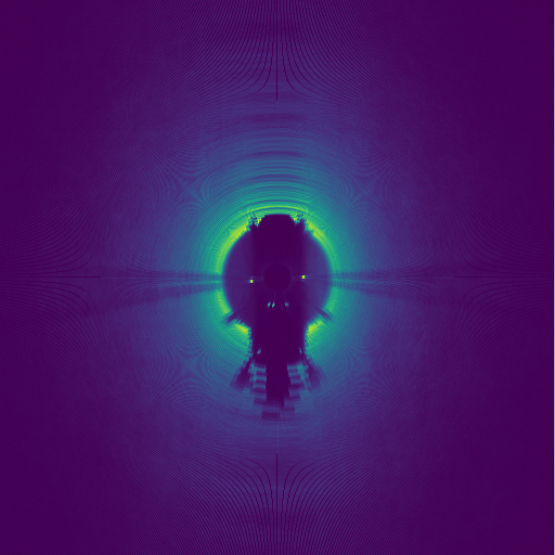} &
		\includegraphics[width=0.9\hsize]{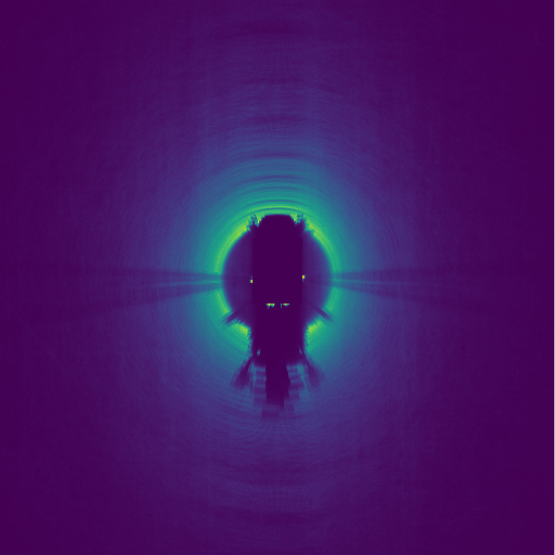} \\
		LiDM & LiDM w/ APE & Dataset 
	\end{tabularx}
	\caption{\textbf{Distribution of point clouds in bird's eye view.} We calculated the marginal distribution of 1,000 random samples  generated by LiDM~\cite{ran2024towards}. With APE, the distribution gets closer to the dataset.}
	\label{fig:lidm_improvement}
\end{figure}

\myparagraph{LiDM improvements.}
We found that LiDM~\cite{ran2024towards} lacks absolute positional bias in the horizontal direction, leading to random azimuth rotation in the \textit{unconditionally} generated samples (see \cref{fig:lidm_improvement} for the visualization).
To ensure a fair comparison, we incorporate an absolute positional embedding (APE) into the diffusion model, similar to recent ViT-based diffusion models~\cite{peebles2023scalable,bao2023all} and ours.
As APE, we add learnable biases $\mathbb{R}^{256\times16\times128}$ after the first convolution layer and re-train the latent diffusion model.
As shown in \cref{fig:lidm_improvement}, the inclusion of APE improved the spatial alignment of the LiDM samples.

\myparagraph{Evaluation metrics.}
Following the related work, we evaluate the distributional similarity between real and generated samples across multiple levels of data representation. We use seven evaluation metrics: Fr\'echet range distance (FRD)\cite{zyrianov2022learning}, Fr\'echet range image distance (FRID)\cite{ran2024towards}, Fr\'echet point cloud distance (FPD)\cite{shu20193d}, Fr\'echet point-based volume distance (FPVD)\cite{ran2024towards}, Fr\'echet sparse volume distance (FSVD)\cite{ran2024towards}, Jensen–Shannon divergence (JSD)\cite{zyrianov2022learning}, and maximum mean discrepancy (MMD)~\cite{zyrianov2022learning}, based on five types of data representations: range images, reflectance images, point clouds, voxels, and bird’s-eye views (BEV). The point cloud, voxel, and BEV representations are derived from the range image. \Cref{tab:fidelity_and_diversity} summarizes the correspondence between the metrics and the representations. Note that only FRD incorporates both range and reflectance modalities, while the other metrics rely solely on the range modality. For each method, we generate 10,000 samples and evaluate them against the entire dataset of real samples, following standard practices in generative models.

\myparagraph{Implementation details.}
All models are implemented using PyTorch.
Training and evaluations were performed on four NVIDIA RTX 6000 Ada GPUs.
We performed a distributed training with automatic mixed precision (AMP).
We used \texttt{torchdiffeq}~\cite{chen2018torchdiffeq} for solving ODEs.
For training 2-RF, we sampled 1M pairs by the \texttt{dopri5} sampler (adaptive step-size) with absolute/relative tolerance of \texttt{1e-5}.
For timestep distillation, we sampled 100k pairs with the same sampler.
All evaluation results are produced by the \texttt{euler} sampler (fixed step-size) for fair comparison.
Our code and pretrained weights are available at \url{https://github.com/kazuto1011/r2flow}.

\subsection{Results}
\label{sec:quantitative_results}

\begin{figure}[t]
	\centering
	\scriptsize
	\definecolor{ours}{rgb}{0.282,0.471,0.816}
	\includegraphics[width=0.8\hsize]{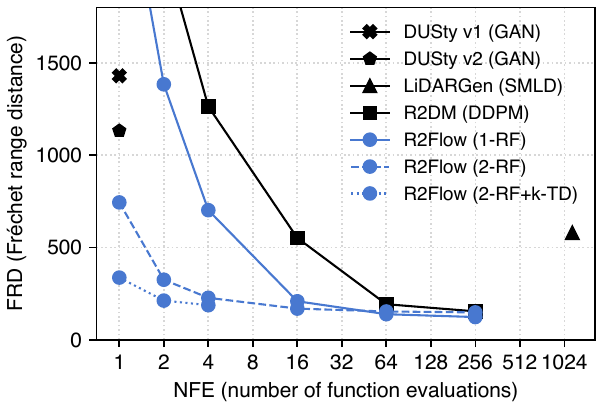}
	\caption{\textbf{Speed--quality tradeoff in FRD.} We compare the methods~\cite{nakashima2021learning,nakashima2023generative,zyrianov2022learning,nakashima2024lidar} that support both range and reflectance modalities. Our R2Flow (\textcolor{ours}{blue lines}) shows the better tradeoff against the baselines (\textcolor{black}{black lines}).}
	\label{fig:frd_tradeoff}
\end{figure}

\myparagraph{Quantitative results.}
\cref{tab:fidelity_and_diversity} shows the evaluation results in three groups from top to bottom
In the first group which compares GAN-based methods, DUSty v2 that generates both range and reflectance images performed well across multiple metrics.
The second group compares the results of the iterative models including ours, with a higher NFE (number of function evaluations).
NFE counts the number of times running neural networks.
With the introduction of APE, LiDM demonstrates improvements across all metrics, particularly in BEV-based metrics, JSD and MMD.
Our R2Flow achieved results comparable to another pixel-space model, R2DM.
On the other hand, the 2-RF scores were slightly lower, which is because the quality upper bound for 2-RF is limited by the parent model 1-RF rather than real data.
Overall, the iterative models are better than the GANs.
The third group compares the results with a fewer NFE. 
While all baseline methods exhibit significant performance degradation, our R2Flow mitigates the degradation through reflow and distillation.
By incorporating the few-step distillation (2-TD and 4-TD), some metrics exhibit results comparable to those achieved with a larger number of steps.
\cref{fig:frd_tradeoff} shows FRD scores as a function of NFEs in details. R2Flow shows a better computational tradeoff.

\begin{table*}[t]
	\definecolor{depth}{rgb}{0.282,0.471,0.816}
	\definecolor{rflct}{rgb}{0.933,0.522,0.290}
	\definecolor{point}{rgb}{0.416,0.800,0.392}
	\definecolor{voxel}{rgb}{0.839,0.373,0.373}
	\definecolor{bev}{rgb}{0.584,0.424,0.706}
	\newcommand{\best}[1]{\cellcolor{gray!20}\textbf{#1}}
	\newcommand{\subopt}[1]{\cellcolor{gray!20}#1}
	\centering
	\scriptsize
	\begin{threeparttable}
		\caption{Quantitative Evaluation of Unconditional Generation on KITTI-360}
		\label{tab:fidelity_and_diversity}
		\begin{tabularx}{\hsize}{lclc CCCCCCC}
			\toprule
			Method                              & Output & Framework       & NFE~$\downarrow$   &
			\twocolorsquare{depth}{rflct}~FRD~$\downarrow$     &
			\twocolorsquare{depth}{depth}~FRID~$\downarrow$ &
			\twocolorsquare{point}{point}~FPD~$\downarrow$     &
			\twocolorsquare{point}{voxel}~FPVD~$\downarrow$ &
			\twocolorsquare{voxel}{voxel}~FSVD~$\downarrow$  &
			\twocolorsquare{bev}{bev}~JSD~$\downarrow$ &
			\twocolorsquare{bev}{bev}~MMD~$\downarrow$ \\
			\midrule
			Vanilla GAN~\cite{caccia2019deep}       & \twocolorsquare{depth}{depth} & GAN         & \s\s\s1 & -                 & 221.89            & 1258.86             & 268.16           & 358.28           & 10.09           & \s\s7.22                       \\
			                                        & \twocolorsquare{depth}{rflct} & GAN         & \s\s\s1 & 2766.43           & 216.45            & 1367.85             & 196.35           & 257.98           & \s8.73          & \s\s5.22                       \\
			DUSty v1~\cite{nakashima2021learning}   & \twocolorsquare{depth}{depth} & GAN         & \s\s\s1 & -                 & \subopt{174.74}   & \s\best{102.11}     & 100.91           & \subopt{118.23}  & \s\subopt{3.90} & \s\s\subopt{2.19}              \\
			                                        & \twocolorsquare{depth}{rflct} & GAN         & \s\s\s1 & \subopt{1429.96}  & 194.25            & \s\subopt{318.06}   & 134.24           & 171.41           & \s5.40          & \s\s2.95                       \\
			DUSty v2~\cite{nakashima2023generative} & \twocolorsquare{depth}{depth} & GAN         & \s\s\s1 & -                 & \best{140.23}     & \s713.59            & \s\subopt{96.69} & 120.00           & \s5.17          & \s\s3.22                       \\
			                                        & \twocolorsquare{depth}{rflct} & GAN         & \s\s\s1 & \best{1132.02}    & 180.71            & \s641.12            & \s\best{88.72}   & \best{106.34}    & \s\best{2.98}   & \s\s\best{0.87}                \\
			\midrule
			LiDARGen~\cite{zyrianov2022learning}    & \twocolorsquare{depth}{rflct} & SMLD        & 1160    & \s542.74          & \s66.45           & \s\s90.29           & \s43.56          & \s50.14          & \s4.04          & \s\s1.74                       \\
			LiDM (original)~\cite{ran2024towards}   & \twocolorsquare{depth}{depth} & DDPM        & \s\s50  & {-}               & \s\s9.21          & \s503.74            & \s19.45          & \s18.08          & 17.40           & \s33.15                        \\
			LiDM~\cite{ran2024towards} + APE        & \twocolorsquare{depth}{depth} & DDPM        & \s\s50  & {-}               & \s\s6.21          & \s393.18            & \s15.25          & \s15.87          & \s4.50          & \s\s0.89                       \\
			                                        & \twocolorsquare{depth}{depth} & DDPM        & \s200   & {-}               & \s\s5.75          & \s372.81            & \s\subopt{14.65} & \s15.20          & \s4.30          & \s\s0.67                       \\
			R2DM~\cite{nakashima2024lidar}          & \twocolorsquare{depth}{rflct} & DDPM        & \s256   & \s154.11          & \s\s\best{3.70}   & \s\s\s\best{3.79}   & \s\best{10.90}   & \s\s\best{9.11}  & \s\subopt{2.19} & \s\s0.73                       \\
			\textbf{R2Flow} (ours)                  & \twocolorsquare{depth}{rflct} & 1-RF        & \s256   & \s\best{122.81}   & \s\s\subopt{4.18} & \s\s\s\subopt{9.32} & \s16.51          & \s\subopt{15.07} & \s\best{2.16}   & \s\s\subopt{0.30}              \\
			                                        & \twocolorsquare{depth}{rflct} & 2-RF        & \s256   & \s\subopt{148.09} & \s\s8.64          & \s\s11.06           & \s29.08          & \s29.20          & \s2.24          & \s\s\best{0.29}                \\
			\midrule
			LiDM~\cite{ran2024towards} + APE        & \twocolorsquare{depth}{depth} & DDPM        & \s\s\s1 & {-}               & 191.00            & 1240.25             & 138.43           & 200.10           & 36.14           & 188.41                         \\
			R2DM~\cite{nakashima2024lidar}          & \twocolorsquare{depth}{rflct} & DDPM        & \s\s\s1 & 2981.89           & 237.41            & \s237.40            & 378.92           & 458.69           & 33.48           & \s89.14                        \\
			\textbf{R2Flow} (ours)                  & \twocolorsquare{depth}{rflct} & 1-RF        & \s\s\s1 & 2724.67           & 254.03            & 3967.51             & 374.89           & 398.95           & 33.76           & \s96.90                        \\
			                                        & \twocolorsquare{depth}{rflct} & 2-RF        & \s\s\s1 & \s743.97          & 119.42            & \s\s27.94           & \s90.92          & 107.72           & \s3.86          & \s\s1.11                       \\
			                                        & \twocolorsquare{depth}{rflct} & 2-RF + 1-TD & \s\s\s1 & \s336.62          & \s54.75           & \s\s13.08           & \s64.03          & \s70.15          & \s\best{2.37}   & \s\s0.35                       \\
			                                        & \twocolorsquare{depth}{rflct} & 2-RF + 2-TD & \s\s\s2 & \s\subopt{212.08} & \s\subopt{22.57}  & \s\s\subopt{11.00}  & \s\subopt{45.56} & \s\subopt{48.98} & \s\subopt{2.39} & \s\s\best{0.31}                \\
			                                        & \twocolorsquare{depth}{rflct} & 2-RF + 4-TD & \s\s\s4 & \s\best{187.10}   & \s\best{15.92}    & \s\s\best{10.92}    & \s\best{38.64}   & \s\best{40.75}   & \s2.40          & \s\s\best{0.31}                \\
			\bottomrule
		\end{tabularx}
		\begin{tablenotes}
			\item Notation: \twocolorsquare{depth}{depth} range image, \twocolorsquare{rflct}{rflct} reflectance image, \twocolorsquare{point}{point} point cloud, \twocolorsquare{voxel}{voxel} voxel, \twocolorsquare{bev}{bev} bird's eye view (BEV).
			\item For each group, we highlight the top-1 scores in \best{bold} and the top-2 scores in \colorbox{gray!20}{shaded}. The JSD and MMD scores are multiplied by $10^2$ and $10^4$, respectively.
		\end{tablenotes}
	\end{threeparttable}
\end{table*}

\begin{figure*}[t]
	\centering
	\scriptsize
	\setlength{\tabcolsep}{2pt}
	\begin{tabularx}{\hsize}{CCCCCC}
		\includegraphics[width=\hsize]{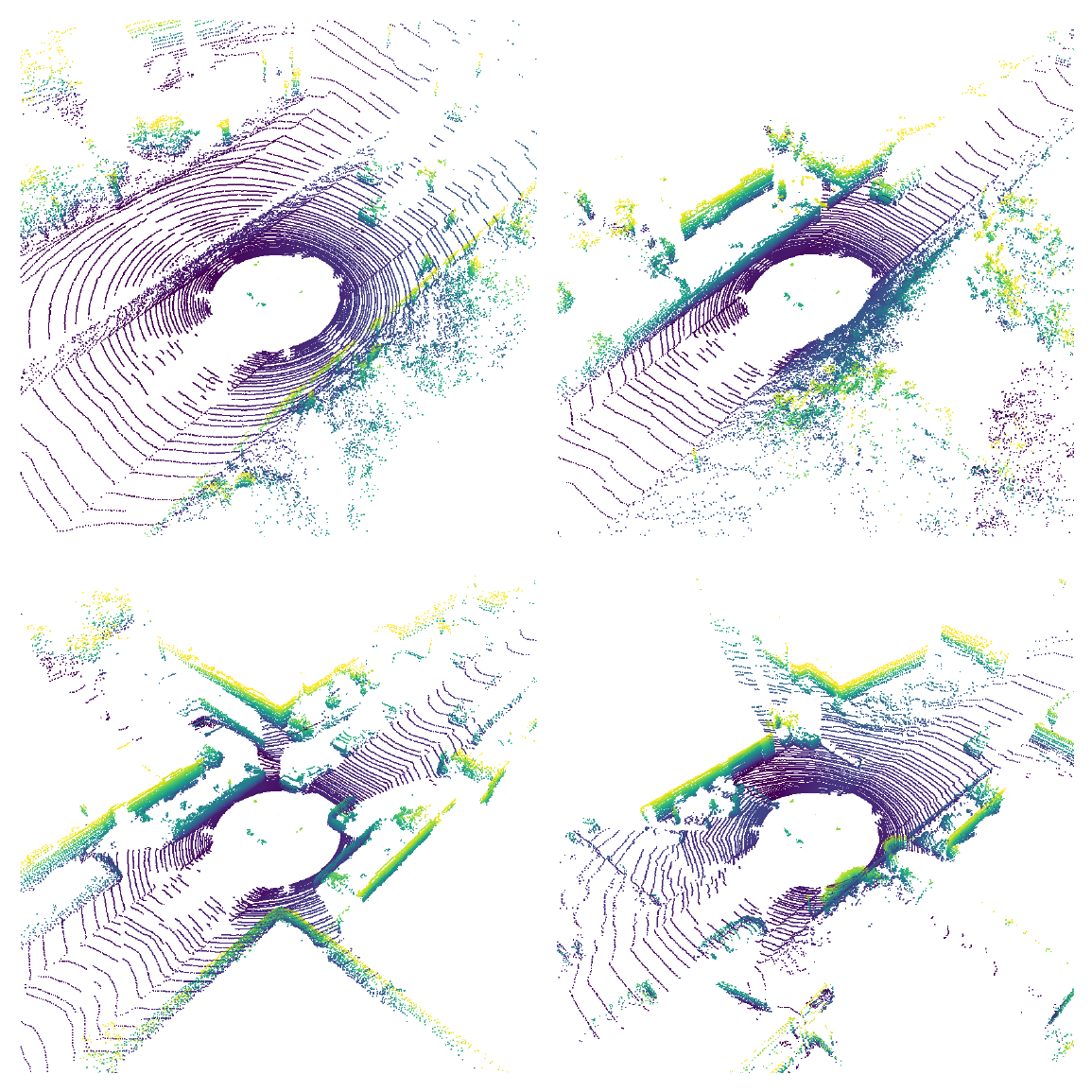} &
		\includegraphics[width=\hsize]{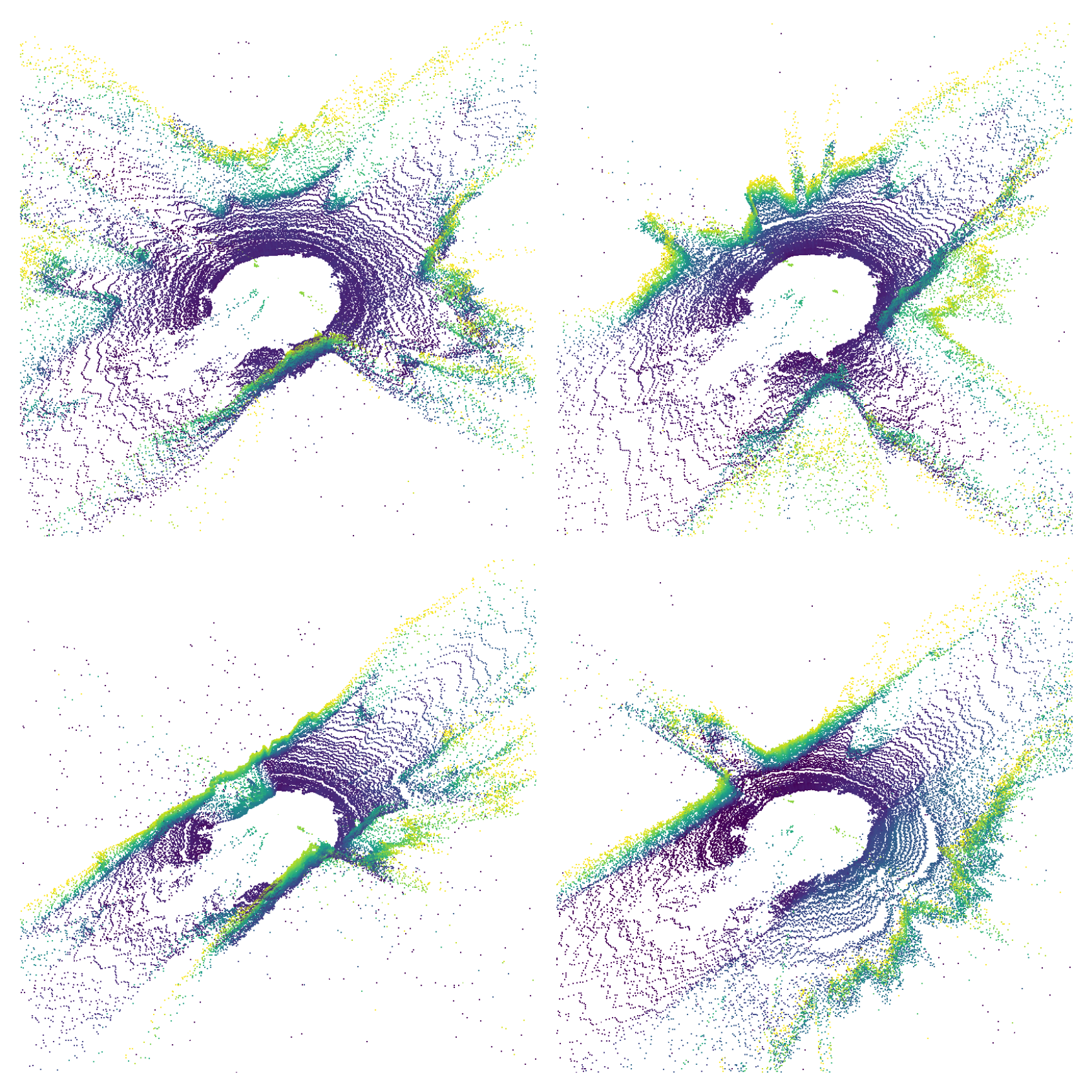} &
		\includegraphics[width=\hsize]{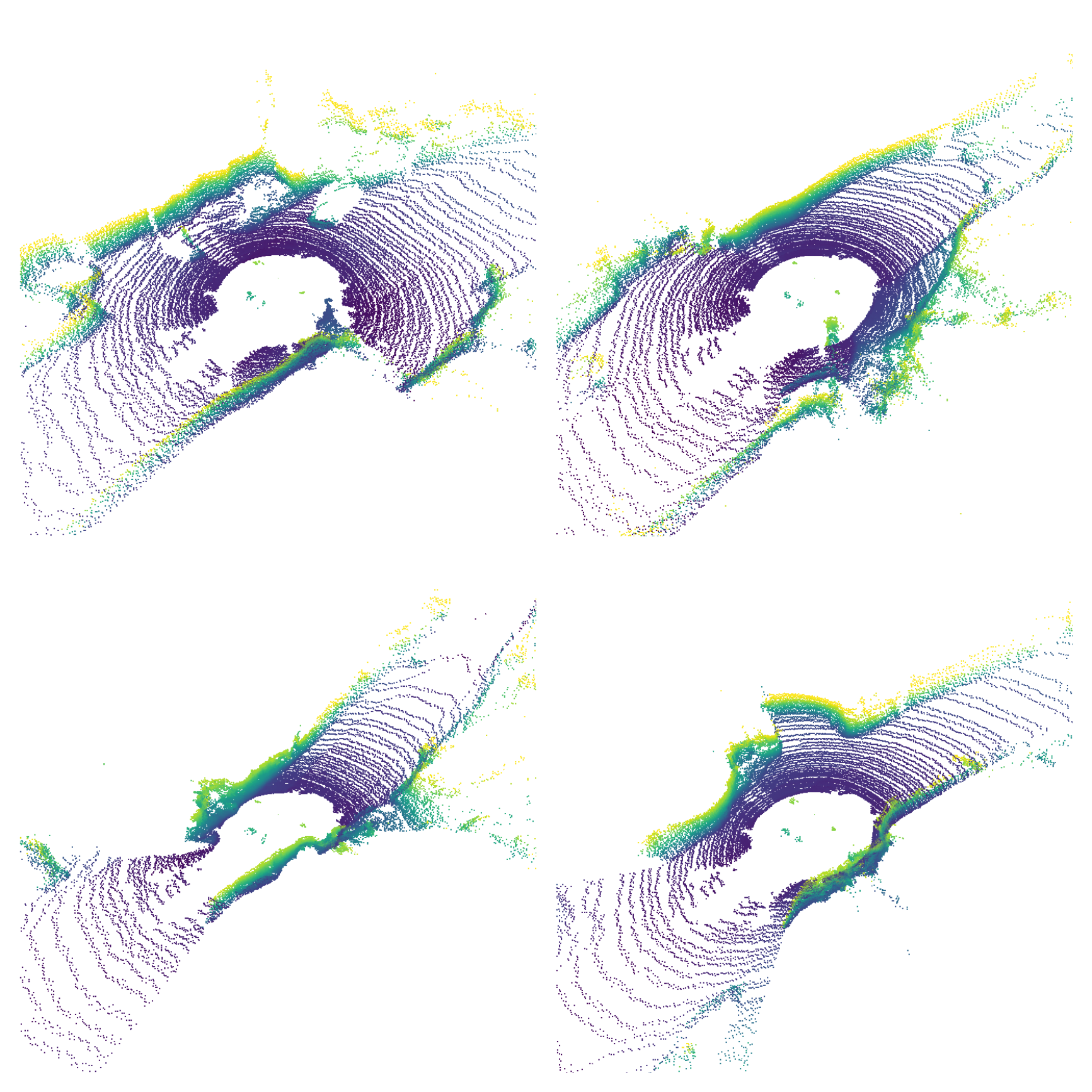} &
		\includegraphics[width=\hsize]{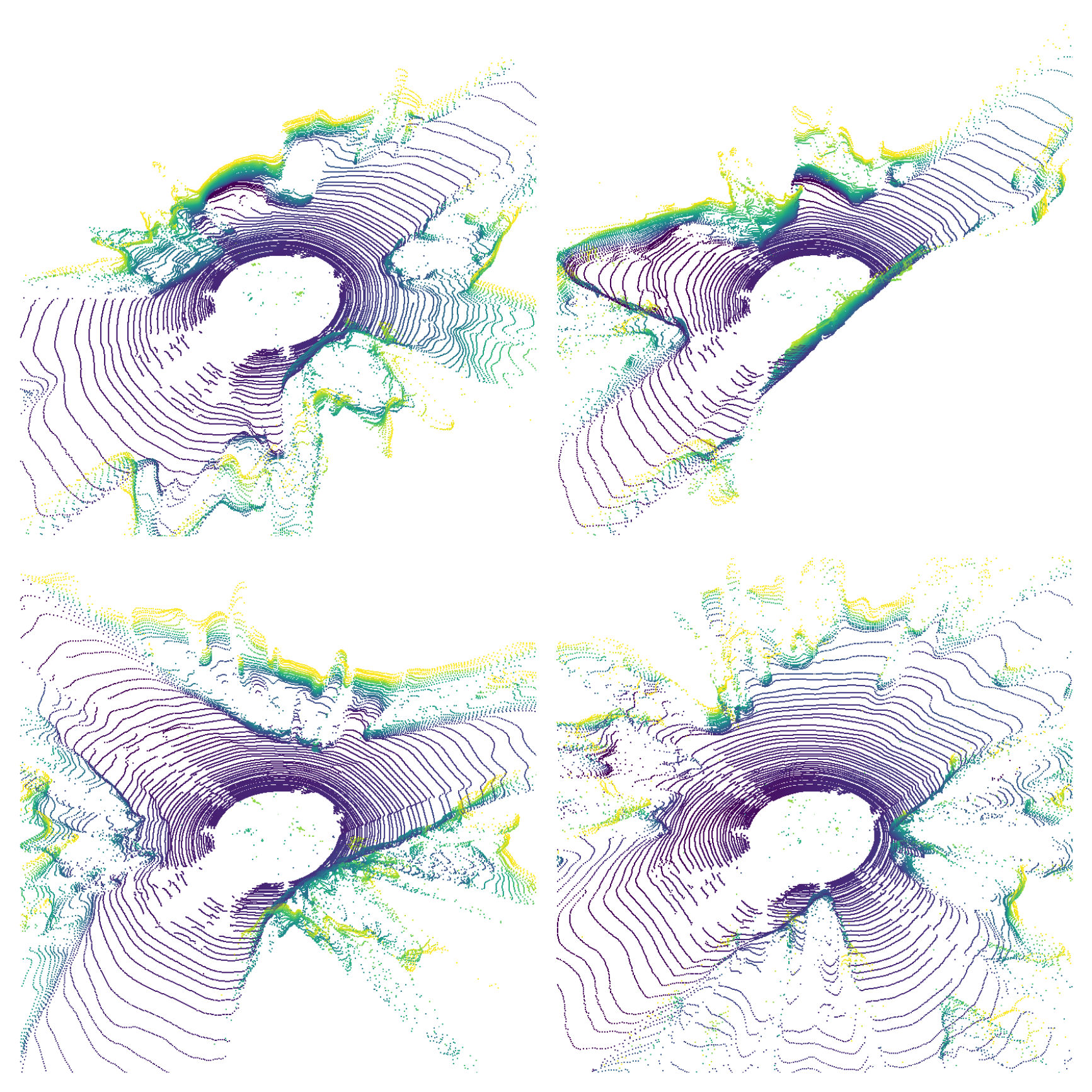} &
		\includegraphics[width=\hsize]{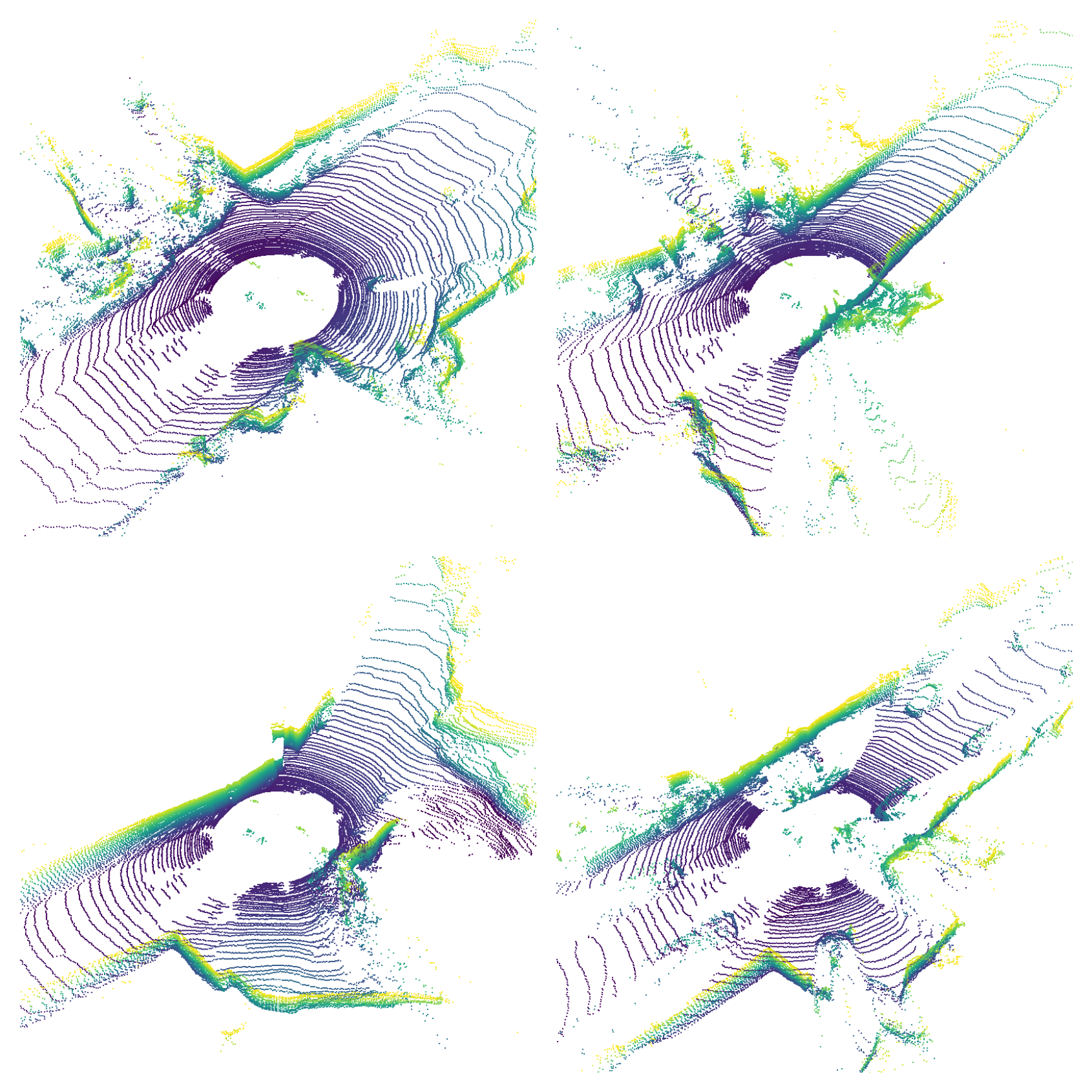} &
		\includegraphics[width=\hsize]{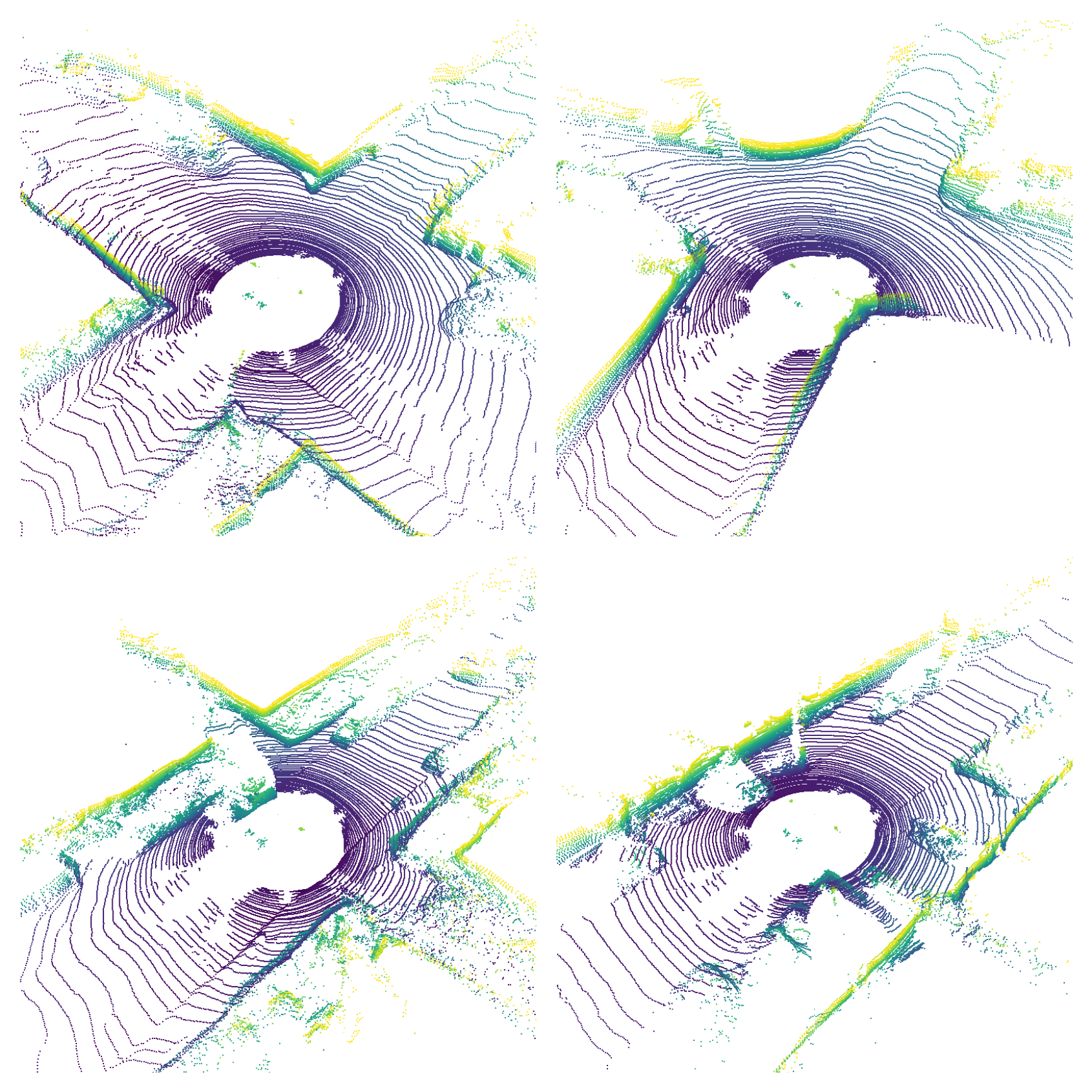} \\
		\includegraphics[width=\hsize]{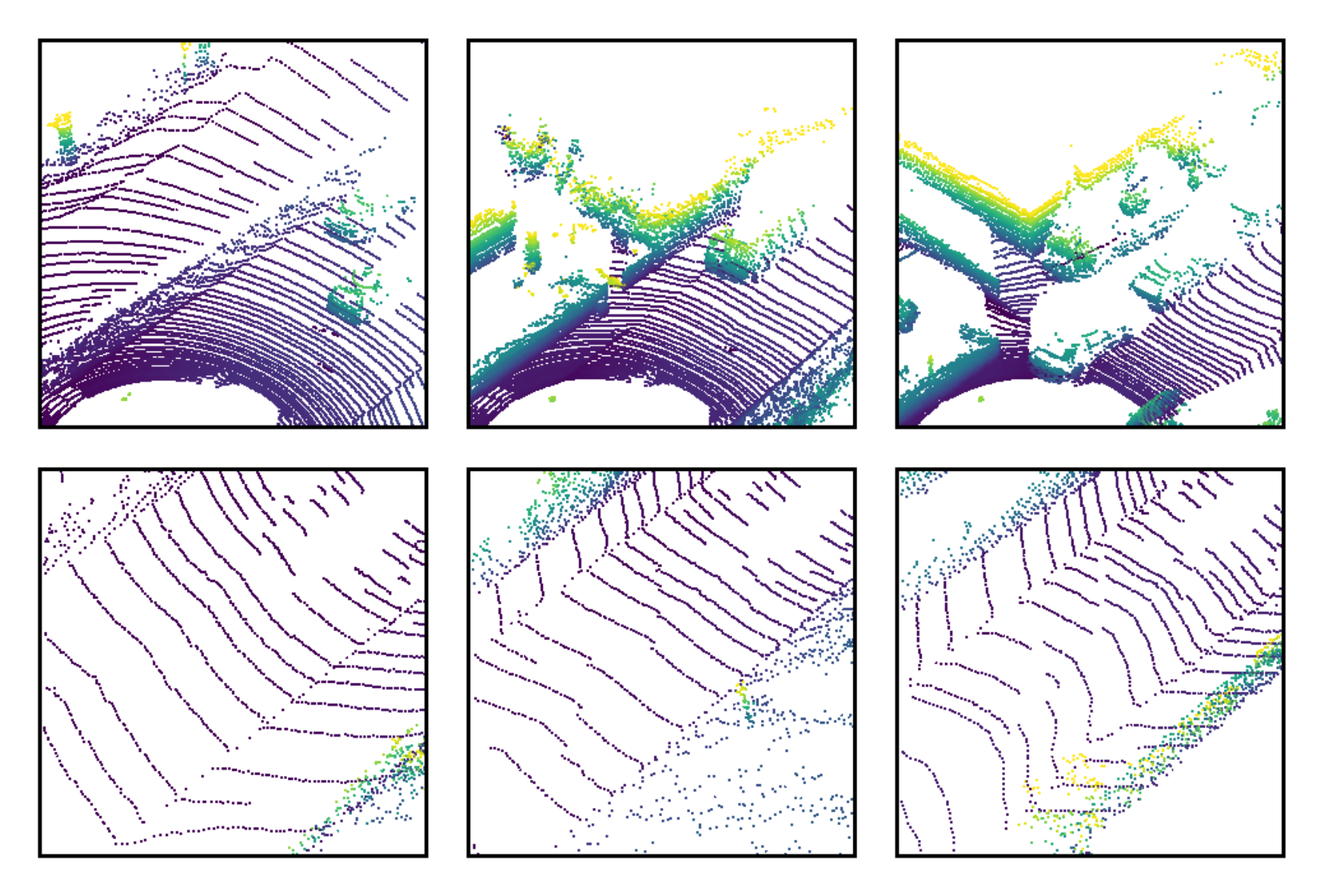} &
		\includegraphics[width=\hsize]{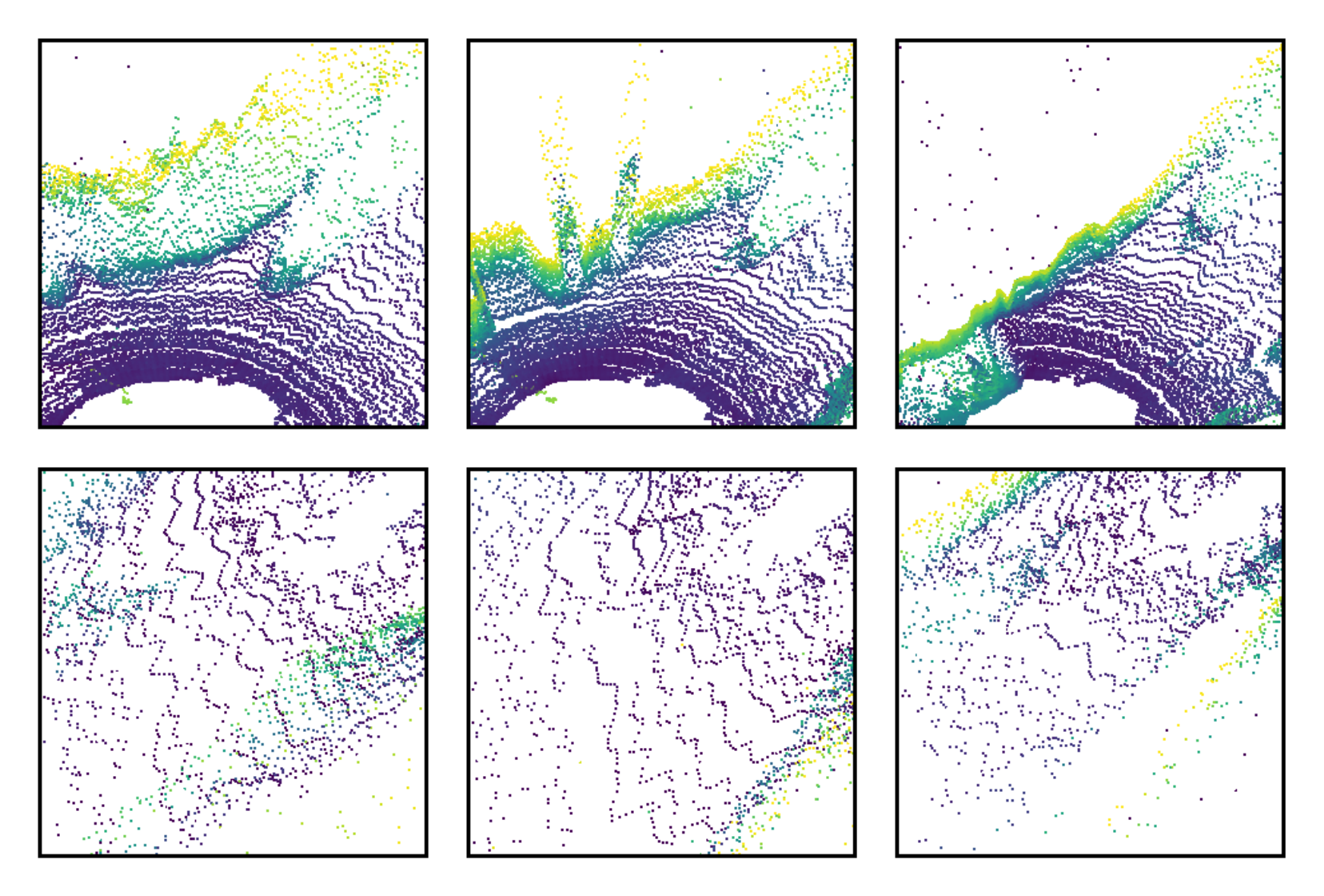} &
		\includegraphics[width=\hsize]{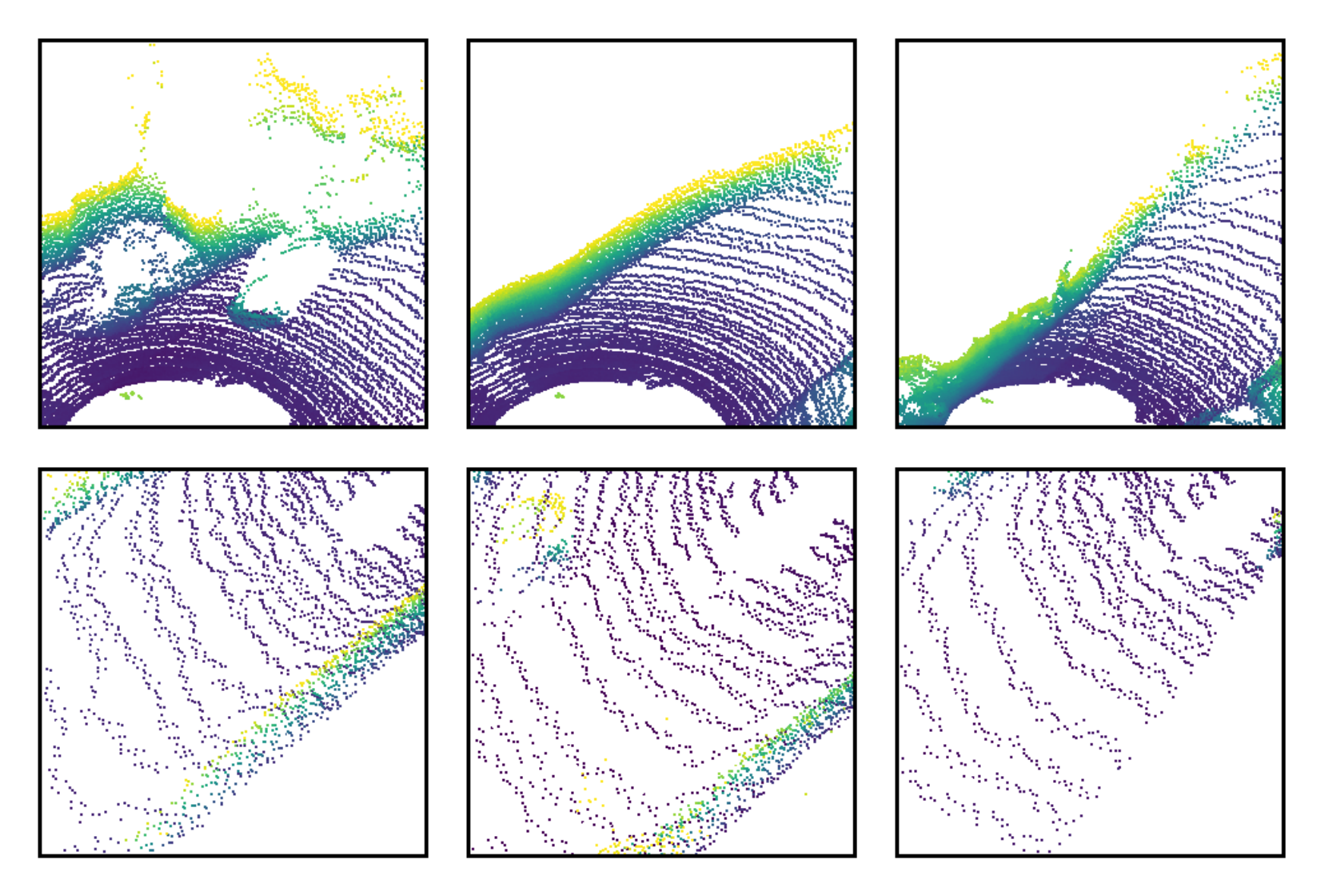} &
		\includegraphics[width=\hsize]{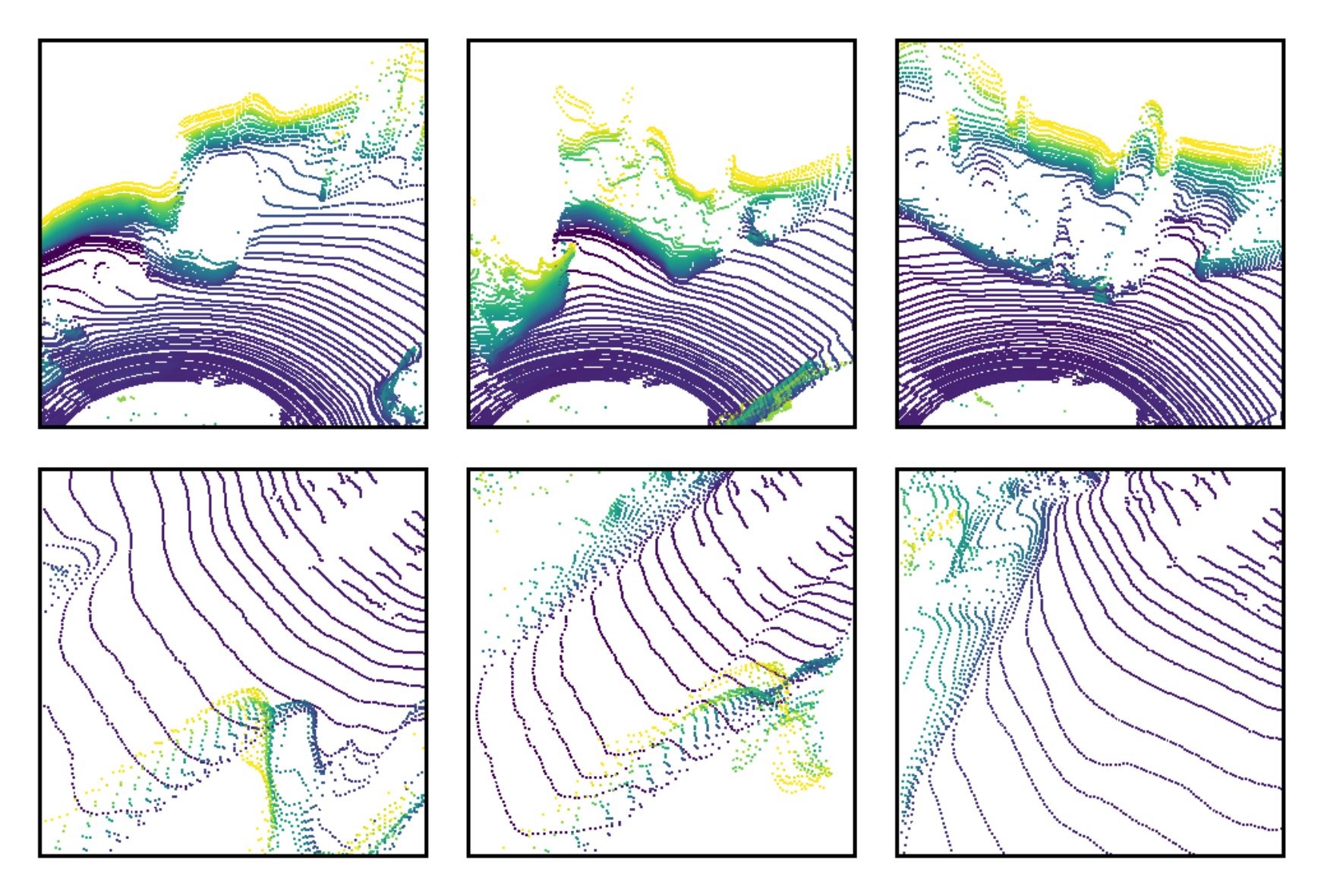} &
		\includegraphics[width=\hsize]{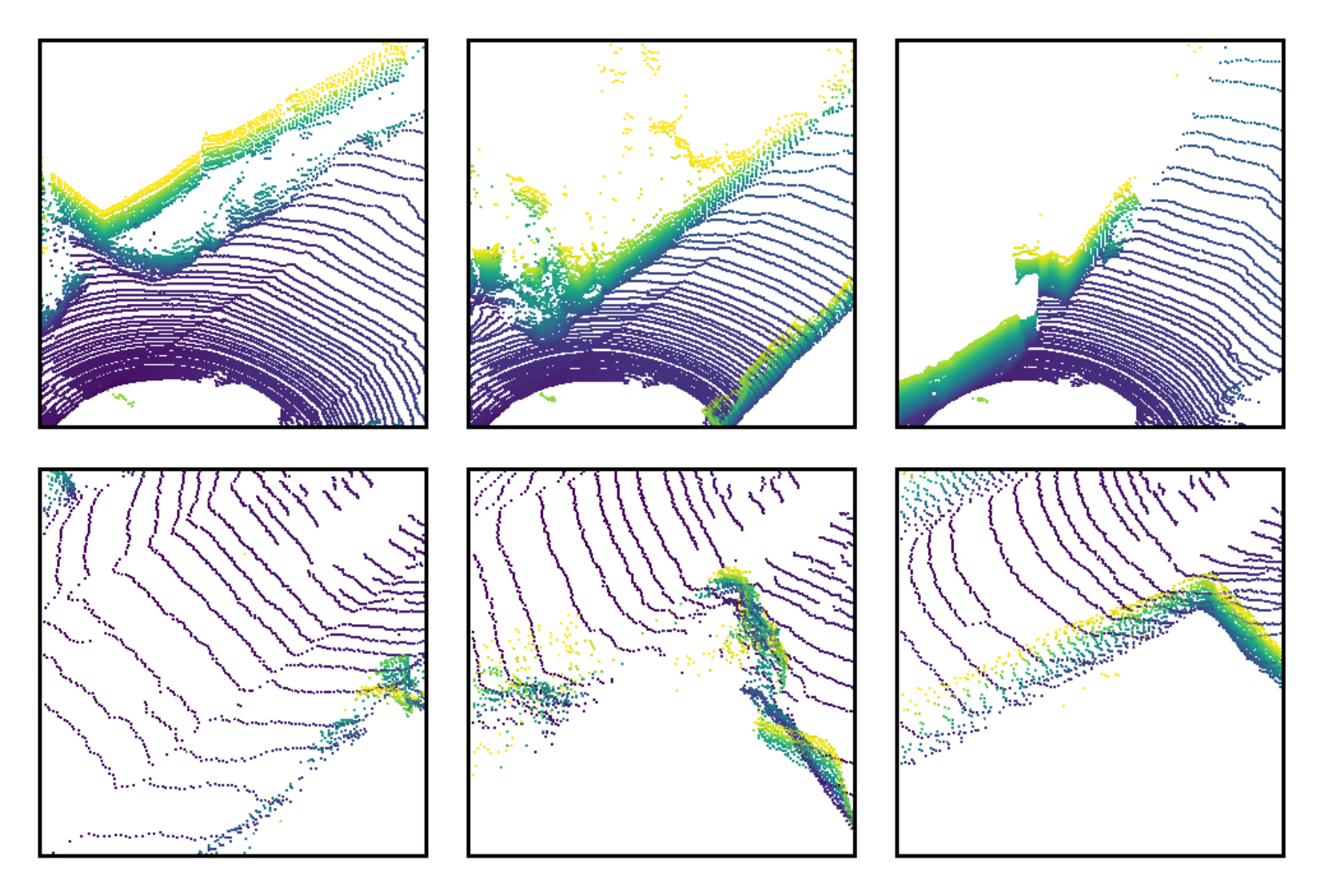} &
		\includegraphics[width=\hsize]{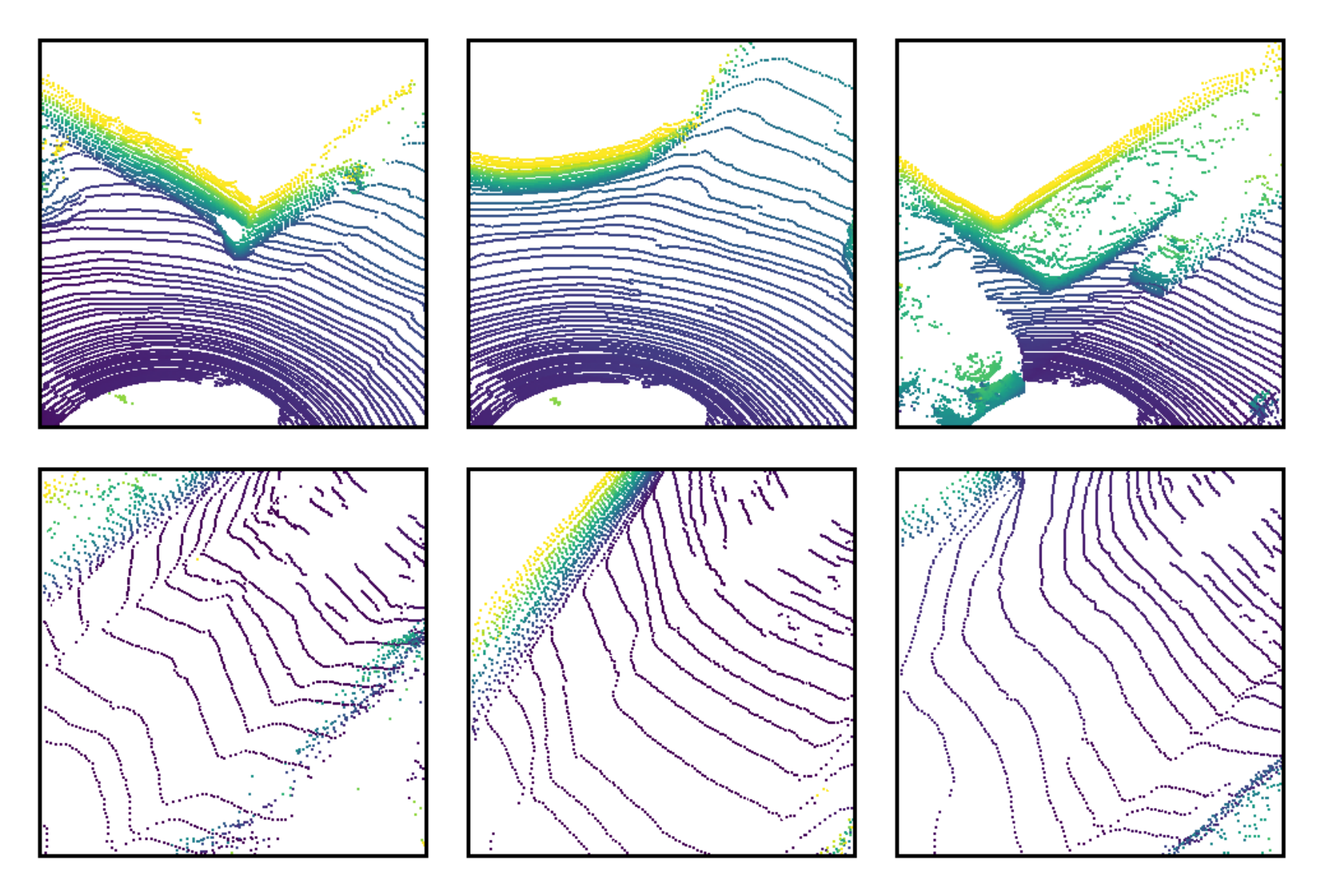} \\
		Training data                    & DUSty v2~\cite{nakashima2023generative} & LiDARGen~\cite{zyrianov2022learning} & LiDM~\cite{ran2024towards} + APE & R2DM~\cite{nakashima2024lidar} & \textbf{R2Flow} (ours) \\
		(KITTI-360~\cite{liao2022kitti-360}) & (GAN, NFE = 1)                          & (SMLD, NFE = 1160)                   & (DDPM, NFE = 200)                & (DDPM, NFE = 256)              & (1-RF, NFE = 256)      
	\end{tabularx}
	\caption{\textbf{Comparison of unconditional generation results.}}
	\label{fig:generated_samples}
\end{figure*}

\myparagraph{Qualitative results.}
In \cref{fig:generated_samples}, we compare LiDAR point clouds generated from DUSty v2~\cite{nakashima2023generative}, LiDARGen~\cite{zyrianov2022learning}, our improved LiDM~\cite{ran2024towards}, R2DM~\cite{nakashima2024lidar}, and our R2Flow.
LiDM, R2DM, and R2Flow demonstrate better quality, such as sharper scan lines and clearer object boundaries, while LiDM exhibits some wavy and blurry boundaries, as previously reported by Ran~\etal~\cite{ran2024towards}.

\myparagraph{Model architecture.}
In \Cref{tab:arch_comparison}, we compare three model architectures for the pixel-space velocity estimator: Efficient U-Net~\cite{saharia2022photorealistic} (CNN) as used in R2DM~\cite{nakashima2024lidar}, ADM U-Net~\cite{dhariwal2021diffusion} (CNN) commonly used for natural images, and HDiT~\cite{crowson2024scalable} (Transformer) as used in ours.
Among the tested configurations, the HDiT-based architecture achieved the best performance.
Although increasing the number of parameters improves the performance of both Efficient U-Net and ADM U-Net, bringing them closer to HDiT, this comes at the cost of higher computational complexity and increased latency.

\begin{table}
	\newcommand{\best}[1]{\textbf{#1}}
	\centering
	\scriptsize
	\begin{threeparttable}
		\caption{Architecture Comparison of Velocity Estimator}
		\label{tab:arch_comparison}
		\begin{tabularx}{\hsize}{lcccC}
			\toprule
			Base architecture                                & FLOPs (G) & Params (M) & Latency (ms) & FRD~\cite{zyrianov2022learning} \\
			\midrule
			Efficient U-Net~\cite{saharia2022photorealistic} & 116.3     & \s31.1     & 15.2         & 151.90                          \\
			~+ larger model size                             & 688.3     & 284.6      & 39.5         & 124.49                          \\
			\midrule
			ADM U-Net~\cite{dhariwal2021diffusion}           & 265.6     & \s87.4     & 24.8         & 140.66                          \\
			~+ larger model size                             & 692.7     & 125.5      & 50.2         & 134.22                          \\
			\midrule
			HDiT~\cite{crowson2024scalable}                  & \s77.8    & \s80.9     & 28.8         & \best{122.81}                   \\
			\bottomrule
		\end{tabularx}
		\begin{tablenotes}
			\item We trained the 1-rectified flow with the different architectures and evaluated FRD with the 256-step Euler sampling.
		\end{tablenotes}
	\end{threeparttable}
\end{table}

\begin{figure}[t]
	\centering
	\footnotesize
	\includegraphics[width=\hsize]{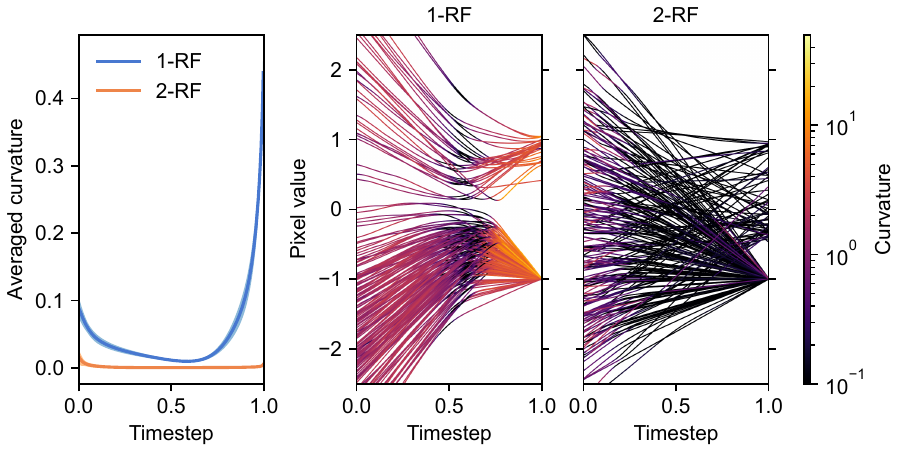}
	\begin{tabularx}{\hsize}{P{30mm}P{35mm}}
		(a) Curvature over time &   
		(b) Trajectories of pixels
	\end{tabularx}
	\caption{\textbf{Trajectory curvature of learned flows}. (a) Trajectory curvature between 1-RF and 2-RF (b) top-200 curved trajectories (0.15\% of all pixels). The pixel value $-1$ at $t=1$ corresponds to \textit{raydrop} noise.}
	\label{fig:straightness}
\end{figure}

\myparagraph{Trajectory curvature.}
To verify the straightening effect by reflow, we measure the trajectory curvature over timestep, defined in prior work~\cite{liu2023flow, lee2024improving}:
\begin{align}
	\label{eq:trajectory_curvature}                                                                                                     
	s\left(t\right)=\| \left(\Phi\left(\bm{x}_0,1\right) - \bm{x}_0\right) - v_\theta\left(\Phi\left(\bm{x}_0,t\right), t\right)\|^2_2, 
\end{align}
where $\bm{x}_0\sim p_0$ and $\Phi\left(\bm{x}_{t_{\mathrm{start}}},t_{\mathrm{end}}\right)$ is the ODE solution from the timestep $t_{\mathrm{start}}$ to $t_{\mathrm{end}}$ with the initial value $\bm{x}_{t_{\mathrm{start}}}$.
As the trajectory becomes straighter,  $s(t)$ approaches zero.
Fig.\ref{fig:straightness}(a) illustrates the trajectory curvature for 1-RF and 2-RF over 256 timesteps.
1-RF exhibits high curvature at both early and late timesteps.
In \cref{fig:straightness}(b), we visualize the top 200 most curved trajectories.
It is evident that many 1-RF trajectories exhibit significant curvature, necessitating numerous sampling steps for generating high-quality samples.
Notably, a part of the trajectories with high curvature near $-1$ correspond to pixels affected by raydrop noise.
The one-time reflow in 2-RF markedly enhances trajectory straightness, as also validated by quantitative results in \cref{tab:fidelity_and_diversity}.

\section{Conclusions}

In this paper, we presented R2Flow, the rectified flow-based generative model for fast and realistic LiDAR data generation.
We verified the effectiveness of our approach in both efficiency and quality in the unconditional generation evaluation.
Future work will focus on exploring the scalability of R2Flow, refining the reflow process to maintain quality, and demonstrating its effectiveness in application tasks such as sparse-to-dense completion, sim-to-real domain adaptation, and anomaly detection.
The trajectory visualization in~\cref{fig:straightness} suggests that the raydrop pixels drifting toward a value of $-1$ may hinder the training of straight flows.
We anticipate that implementing a raydrop-aware architecture~\cite{nakashima2021learning,nakashima2023generative} could mitigate this issue.
 
\bibliographystyle{ieeetr}
\bibliography{main}

\begin{thebibliography}{10}

\bibitem{caccia2019deep}
L.~Caccia, H.~van Hoof, A.~Courville, and J.~Pineau, ``Deep generative modeling of {LiDAR} data,'' in {\em Proceedings of the IEEE/RSJ International Conference on Intelligent Robots and Systems (IROS)}, pp.~5034--5040, 2019.

\bibitem{nakashima2021learning}
K.~Nakashima and R.~Kurazume, ``Learning to drop points for {LiDAR} scan synthesis,'' in {\em Proceedings of the IEEE/RSJ International Conference on Intelligent Robots and Systems (IROS)}, pp.~222--229, 2021.

\bibitem{nakashima2023generative}
K.~Nakashima, Y.~Iwashita, and R.~Kurazume, ``Generative range imaging for learning scene priors of {3D} {LiDAR} data,'' in {\em Proceedings of the IEEE/CVF Winter Conference on Applications of Computer Vision (WACV)}, pp.~1256--1266, 2023.

\bibitem{nakashima2024lidar}
K.~Nakashima and R.~Kurazume, ``{LiDAR} data synthesis with denoising diffusion probabilistic models,'' in {\em Proceedings of the IEEE International Conference on Robotics and Automation (ICRA)}, pp.~14724--14731, 2024.

\bibitem{zyrianov2022learning}
V.~Zyrianov, X.~Zhu, and S.~Wang, ``Learning to generate realistic {LiDAR} point clouds,'' in {\em Proceedings of the European Conference on Computer Vision (ECCV)}, pp.~17--35, 2022.

\bibitem{ran2024towards}
H.~Ran, V.~Guizilini, and Y.~Wang, ``Towards realistic scene generation with {LiDAR} diffusion models,'' in {\em Proceedings of the IEEE/CVF Conference on Computer Vision and Pattern Recognition (CVPR)}, 2024.

\bibitem{xiong2023learning}
Y.~Xiong, W.-C. Ma, J.~Wang, and R.~Urtasun, ``Learning compact representations for lidar completion and generation,'' in {\em Proceedings of the IEEE/CVF Conference on Computer Vision and Pattern Recognition (CVPR)}, pp.~1074--1083, 2023.

\bibitem{hu2024rangeldm}
Q.~Hu, Z.~Zhang, and W.~Hu, ``{RangeLDM}: Fast realistic {LiDAR} point cloud generation,'' in {\em Proceedings of the European Conference on Computer Vision (ECCV)}, p.~115–135, 2024.

\bibitem{bond-taylor2022deep}
S.~Bond-Taylor, A.~Leach, Y.~Long, and C.~G. Willcocks, ``Deep generative modelling: A comparative review of {VAE}s, {GAN}s, normalizing flows, energy-based and autoregressive models,'' {\em IEEE Transactions on Pattern Analysis and Machine Intelligence (TPAMI)}, vol.~44, no.~11, pp.~7327--7347, 2022.

\bibitem{song2021score-based}
Y.~Song, J.~Sohl-Dickstein, D.~P. Kingma, A.~Kumar, S.~Ermon, and B.~Poole, ``Score-based generative modeling through stochastic differential equations,'' in {\em Proceedings of the International Conference on Learning Representations (ICLR)}, 2021.

\bibitem{peebles2023scalable}
W.~Peebles and S.~Xie, ``Scalable diffusion models with transformers,'' in {\em Proceedings of the IEEE/CVF International Conference on Computer Vision (ICCV)}, pp.~4195--4205, 2023.

\bibitem{liu2023flow}
X.~Liu, C.~Gong, and Q.~Liu, ``Flow straight and fast: Learning to generate and transfer data with rectified flow,'' in {\em Proceedings of the International Conference on Learning Representations (ICLR)}, 2023.

\bibitem{lee2024improving}
S.~Lee, Z.~Lin, and G.~Fanti, ``Improving the training of rectified flows,'' in {\em Advances in Neural Information Processing Systems (NeurIPS)}, vol.~37, pp.~63082--63109, 2024.

\bibitem{lipman2023flow}
Y.~Lipman, R.~T.~Q. Chen, H.~Ben-Hamu, M.~Nickel, and M.~Le, ``Flow matching for generative modeling,'' in {\em Proceedings of the International Conference on Learning Representations (ICLR)}, 2023.

\bibitem{tong2024improving}
A.~Tong, K.~FATRAS, N.~Malkin, G.~Huguet, Y.~Zhang, J.~Rector-Brooks, G.~Wolf, and Y.~Bengio, ``Improving and generalizing flow-based generative models with minibatch optimal transport,'' {\em Transactions on Machine Learning Research (TMLR)}, 2024.

\bibitem{crowson2024scalable}
K.~Crowson, S.~A. Baumann, A.~Birch, T.~M. Abraham, D.~Z. Kaplan, and E.~Shippole, ``Scalable high-resolution pixel-space image synthesis with hourglass diffusion transformers,'' in {\em Proceedings of the International Conference on Machine Learning (ICML)}, 2024.

\bibitem{liao2022kitti-360}
Y.~Liao, J.~Xie, and A.~Geiger, ``{KITTI-360}: A novel dataset and benchmarks for urban scene understanding in {2D} and {3D},'' {\em IEEE Transactions on Pattern Analysis and Machine Intelligence (TPAMI)}, vol.~45, no.~3, pp.~3292--3310, 2022.

\bibitem{kingma2014auto-encoding}
D.~P. Kingma and M.~Welling, ``Auto-encoding variational bayes,'' in {\em Proceedings of the International Conference on Learning Representations (ICLR)}, 2014.

\bibitem{vandenoord2017neural}
A.~Van Den~Oord, O.~Vinyals, {\em et~al.}, ``Neural discrete representation learning,'' in {\em Advances in Neural Information Processing Systems (NeurIPS)}, vol.~30, 2017.

\bibitem{goodfellow2014generative}
I.~Goodfellow, J.~Pouget-Abadie, M.~Mirza, B.~Xu, D.~Warde-Farley, S.~Ozair, A.~Courville, and Y.~Bengio, ``Generative adversarial nets,'' in {\em Advances in Neural Information Processing Systems (NeurIPS)}, pp.~2672--2680, 2014.

\bibitem{song2019generative}
Y.~Song and S.~Ermon, ``Generative modeling by estimating gradients of the data distribution,'' in {\em Advances in Neural Information Processing Systems (NeurIPS)}, pp.~11895--11907, 2019.

\bibitem{song2020improved}
Y.~Song and S.~Ermon, ``Improved techniques for training score-based generative models,'' in {\em Advances in Neural Information Processing Systems (NeurIPS)}, vol.~33, pp.~12438--12448, 2020.

\bibitem{ho2020denoising}
J.~Ho, A.~Jain, and P.~Abbeel, ``Denoising diffusion probabilistic models,'' in {\em Advances in Neural Information Processing Systems (NeurIPS)}, vol.~33, pp.~6840--6851, 2020.

\bibitem{kingma2021variational}
D.~Kingma, T.~Salimans, B.~Poole, and J.~Ho, ``Variational diffusion models,'' in {\em Advances in Neural Information Processing Systems (NeurIPS)}, vol.~34, pp.~21696--21707, 2021.

\bibitem{rombach2022high-resolution}
R.~Rombach, A.~Blattmann, D.~Lorenz, P.~Esser, and B.~Ommer, ``High-resolution image synthesis with latent diffusion models,'' in {\em Proceedings of the IEEE/CVF Conference on Computer Vision and Pattern Recognition (CVPR)}, pp.~10684--10695, 2022.

\bibitem{hassani2023neighborhood}
A.~Hassani, S.~Walton, J.~Li, S.~Li, and H.~Shi, ``Neighborhood attention transformer,'' in {\em Proceedings of the IEEE/CVF Conference on Computer Vision and Pattern Recognition (CVPR)}, pp.~6185--6194, 2023.

\bibitem{dosovitskiy2021image}
A.~Dosovitskiy, L.~Beyer, A.~Kolesnikov, D.~Weissenborn, X.~Zhai, T.~Unterthiner, M.~Dehghani, M.~Minderer, G.~Heigold, S.~Gelly, J.~Uszkoreit, and N.~Houlsby, ``An image is worth 16x16 words: Transformers for image recognition at scale,'' in {\em Proceedings of the International Conference on Learning Representations (ICLR)}, 2021.

\bibitem{saharia2022photorealistic}
C.~Saharia, W.~Chan, S.~Saxena, L.~Li, J.~Whang, E.~L. Denton, K.~Ghasemipour, R.~Gontijo~Lopes, B.~Karagol~Ayan, T.~Salimans, {\em et~al.}, ``Photorealistic text-to-image diffusion models with deep language understanding,'' in {\em Advances in Neural Information Processing Systems (NeurIPS)}, vol.~35, pp.~36479--36494, 2022.

\bibitem{yang2024tulip}
B.~Yang, P.~Pfreundschuh, R.~Siegwart, M.~Hutter, P.~Moghadam, and V.~Patil, ``{TULIP}: Transformer for upsampling of {LiDAR} point clouds,'' in {\em Proceedings of the IEEE/CVF Conference on Computer Vision and Pattern Recognition (CVPR)}, pp.~15354--15364, 2024.

\bibitem{su2021roformer}
J.~Su, Y.~Lu, S.~Pan, B.~Wen, and Y.~Liu, ``{RoFormer}: Enhanced transformer with rotary position embedding,'' {\em arXiv:2104.09864}, 2021.

\bibitem{bao2023all}
F.~Bao, S.~Nie, K.~Xue, Y.~Cao, C.~Li, H.~Su, and J.~Zhu, ``All are worth words: A {ViT} backbone for diffusion models,'' in {\em Proceedings of the IEEE/CVF Conference on Computer Vision and Pattern Recognition (CVPR)}, pp.~22669--22679, 2023.

\bibitem{shu20193d}
D.~W. Shu, S.~W. Park, and J.~Kwon, ``{3D} point cloud generative adversarial network based on tree structured graph convolutions,'' in {\em Proceedings of the IEEE/CVF International Conference on Computer Vision (ICCV)}, pp.~3859--3868, 2019.

\bibitem{chen2018torchdiffeq}
R.~T.~Q. Chen, ``torchdiffeq,'' 2018.

\bibitem{dhariwal2021diffusion}
P.~Dhariwal and A.~Nichol, ``Diffusion models beat gans on image synthesis,'' in {\em Advances in Neural Information Processing Systems (NeurIPS)}, vol.~34, pp.~8780--8794, 2021.

\end{thebibliography}
	
\end{document}